\documentclass[sigconf]{acmart}

\usepackage{booktabs} 
\usepackage{hyperref}
\usepackage[utf8]{inputenc} 
\usepackage{url}
\usepackage{amsfonts} 
\usepackage{nicefrac} 

\usepackage{graphicx}
\usepackage{subcaption}
\usepackage{algorithm}
\usepackage{algorithmic}
\usepackage{empheq}
\usepackage{enumitem}
\usepackage{color}
\usepackage{mdframed}
\usepackage{Definitions}
\graphicspath{{./Figs/}}

\setcopyright{rightsretained}

\setcopyright{rightsretained}
\acmDOI{10.475/123_4}
\acmISBN{123-4567-24-567/08/06}
\acmConference[KDD'17]{ACM conference}{August 2017}{Halifax, Canada} 
\acmYear{2017}
\copyrightyear{2017}
\acmPrice{15.00}

\begin{document}
\title{Deep Coevolutionary Network: Embedding User and Item Features for Recommendation}

\author{Hanjun Dai*}
\affiliation{%
  \institution{Georgia Institute of Technology}
}
\email{hanjundai@gatech.edu}

\author{Yichen Wang*}
\affiliation{%
  \institution{Georgia Institute of Technology}
}
\email{yichen.wang@gatech.edu}

\author{Rakshit Trivedi}
\affiliation{%
  \institution{Georgia Institute of Technology}
}
\email{rstrivedi@gatech.edu}

\author{Le Song}
\affiliation{
  \institution{Georgia Institute of Technology}
}
\email{lsong@cc.gatech.edu}


\begin{abstract}
Recommender systems often use latent features to explain the behaviors of users and capture the properties of items. As users interact with different items over time, user and item features can influence each other, evolve and co-evolve over time. The compatibility of user and item's feature further influence the future interaction between users and items. 

Recently, point process based models have been proposed in the literature aiming to capture the temporally evolving nature of these latent features. However, these models often make strong parametric assumptions about the evolution process of the user and item latent features, which may not reflect the reality, and has limited power in expressing the complex and nonlinear dynamics underlying these processes. 

To address these limitations, we propose a novel deep coevolutionary network model (DeepCoevolve), for learning user and item features based on their interaction graph. DeepCoevolve use recurrent neural network (RNN) over evolving networks to define the intensity function in point processes, which allows the model to capture complex mutual influence between users and items, and the feature evolution over time. We also develop an efficient procedure for training the model parameters, and show that the learned models lead to significant improvements in recommendation and activity prediction compared to previous state-of-the-arts parametric models. 
\end{abstract}

%
%
%



\maketitle

\section{Introduction}

\setlength{\abovedisplayskip}{4pt}
\setlength{\abovedisplayshortskip}{1pt}
\setlength{\belowdisplayskip}{4pt}
\setlength{\belowdisplayshortskip}{1pt}
\setlength{\jot}{2pt}
\setlength{\floatsep}{2ex}
\setlength{\textfloatsep}{2ex}

Making proper recommendation of items to users at the right time is a fundamental task in e-commerce platforms and social service websites. The compatibility between user's interest and item's property is a good predictive factor on whether the user will interact with the item in future. Conversely, the interactions between users and items further drives the evolution of user interests and item features. 
As users interact with different items, users' interests and items' features can also {co-evolve} over time,~\ie, their features are intertwined and can influence each other: 
\begin{itemize}[leftmargin=*,nosep,nolistsep]
\vspace{-1mm}
  \item From \emph{User} to \emph{item}. In discussion forums such as Reddit, although a group (item) is initially created for statistics topics, users with very different interest profiles can join this group. Hence, the participants can shape the features of the group through their postings. It is likely that this group can finally become one about deep learning if most users discuss about deep learning. 
  \item From \emph{Item} to \emph{user}. As the group is evolving towards topics on deep learning, some users may become more interested in such topics, and they may participate in other specialized groups. On the contrary, some users may gradually gain interests in math groups, lose interests in statistics  group. 
\end{itemize}

Such coevolutionary nature of user and item features raises very interesting and challenging questions: How to model coevolutionary features? How to efficiently train such models on large scale data?
There are previous attempts which divide time into epochs, and perform tensor factorization to learn the latent features~\citep{ChiKol12,Koren09,YanLonSmoEtal11}. These methods are not able to capture the fine grained temporal dynamics of user-item interactions, and can not answer the query related to time of interaction. Recent, point processes which treat event times as random variables have emerged as a good framework for modeling such temporal feature evolution process~\citep{DuWanHeetal15, WanDuTriSon16}. However, these previous work make strong parametric assumptions about the functional form of the generative processes, which may not reflect the reality, and is not accurate enough to capture the complex and \emph{nonlinear} co-evolution of user and item features in real world.

\begin{figure*}[t]
\small
\centering
\begin{tabular}{ccc}
  \includegraphics[width=0.32\textwidth]{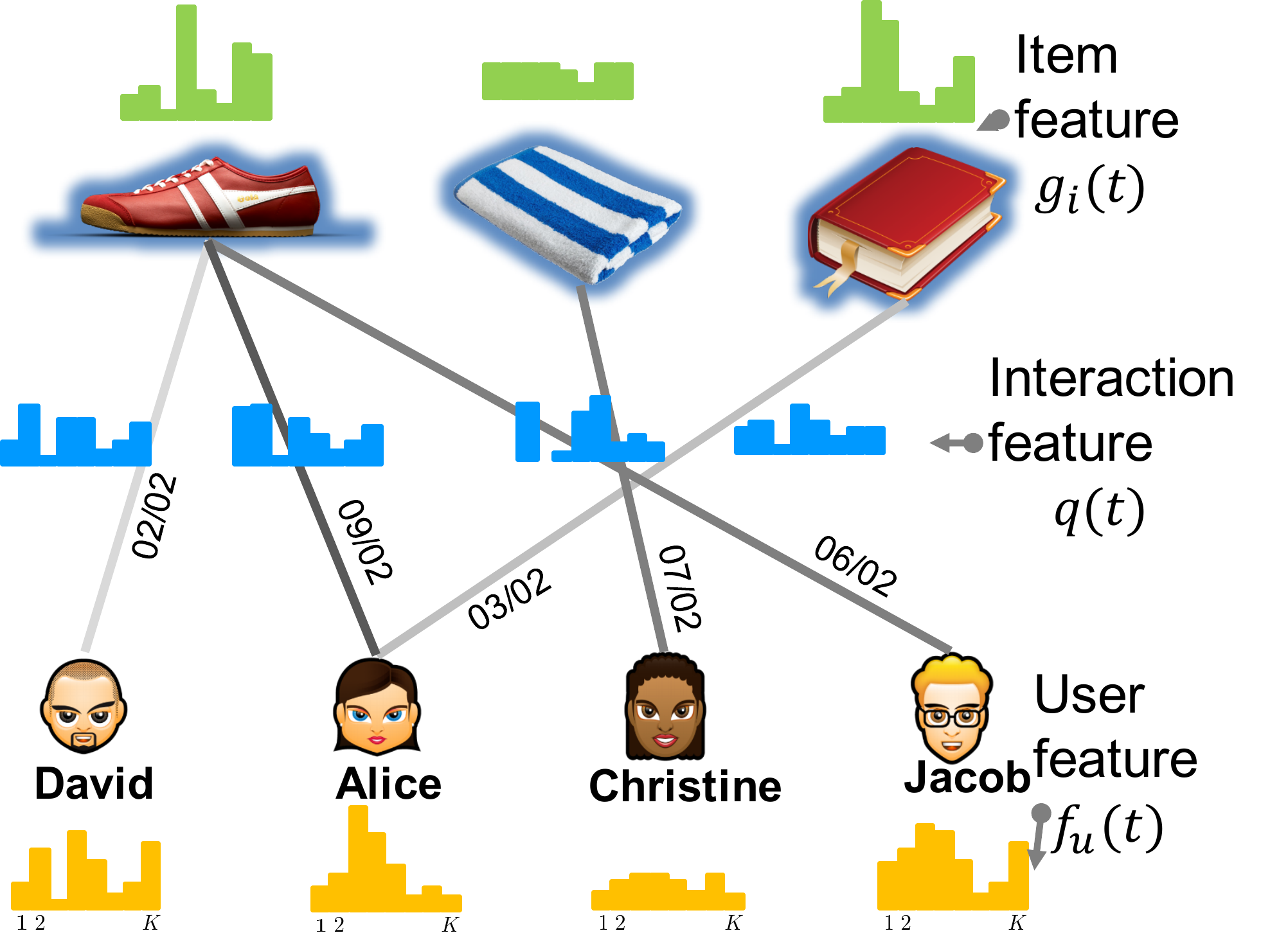}
  &~~~~~&
  \includegraphics[width=0.32\textwidth]{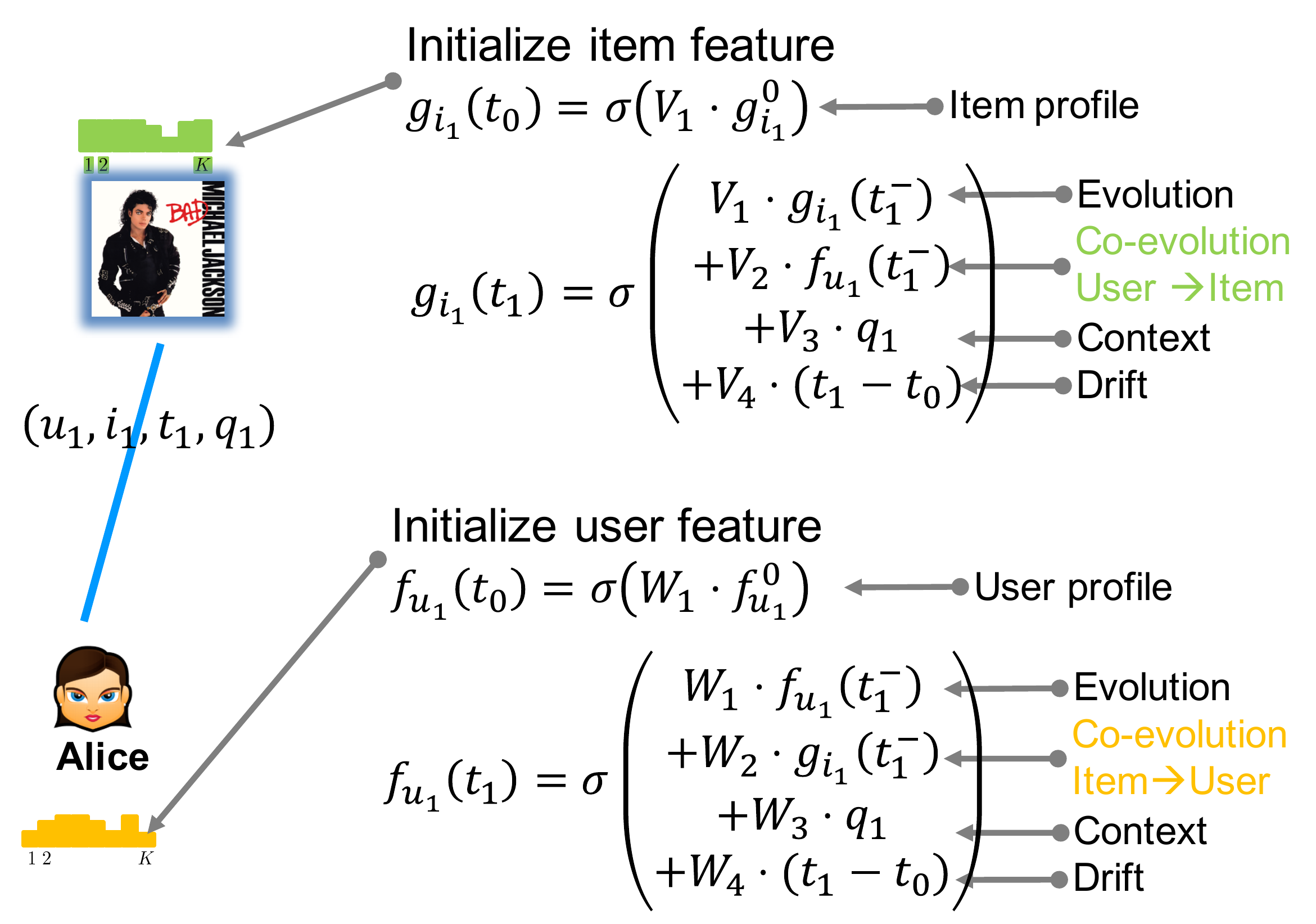} \\
  (a) &~~~~~& (b)
\vspace{-4mm}
\end{tabular}
\caption{Model illustration. (a) User-item interaction as evolving bipartite graph. Each edge stands for a tuple and contains the information of user, item, interaction time, and interaction feature. (b) Deep coevolutionary feature embedding processes. The embeddings of users and items are updated at each event time, by a nonlinear activation function $\sigma(\cdot)$ and four terms: self evolution, co-evolution, context (interaction feature), and self drift.}
  \label{fig:demo}
\vspace{-2mm}
\end{figure*}

To address the limitation in previous point process based methods, we propose a novel deep coevolutionary network model (DeepCoevolve) which defines point process intensities using recurrent neural network (RNN) over evolving interaction networks. Our model can generate an effective representation/embedding of the underlying user and item latent feature without assuming a fixed parametric forms in advance. Figure~\ref{fig:demo} summarizes our framework. In particular, our work makes the following contributions:
 
\begin{itemize}[leftmargin=*,nosep,nolistsep]
	\item {\bf Novel model.} We propose a novel deep learning model that captures the \emph{nonlinear} coevolution nature of users' and items' embeddings in a nonparametric way. It assigns an evolving feature embedding process for each user and item, and the co-evolution of these latent feature processes is modeled with two parallel components: (i) \emph{item} $\to$ \emph{user} component, a user's latent feature is determined by the nonlinear embedding of latent features of the items he interacted with; and (ii) \emph{user} $\to$ \emph{item} component, an item's latent features are also determined by the latent features of the users who interact with the item.
	
	\item {\bf Efficient Training.} We use RNN to parametrize the {interdependent} and intertwined user and item embeddings. The increased flexibility and generality further introduces technical challenges on how to train RNN on the \emph{evolving networks}. The co-evolution nature of the model makes the samples inter-dependent and not identically distributed, which is significantly more challenging than the typical assumptions in training deep models. We propose an efficient stochastic training algorithm that makes the BTPP tractable in the co-evolving network. 
	
	\item {\bf Strong performance.} We evaluate our method over multiple datasets, verifying that our method leads to significant improvements in user behavior prediction compared to state-of-the-arts. It further verifies the importance of using nonparametric point process in recommendation systems. Precise time prediction is especially novel and not possible by most prior work. 	
\end{itemize}

\vspace{-1.5mm}
\section{Background on Temporal Point Processes}
A temporal point process~\citep{CoxIsh1980, CoxLew2006, AalBorGje08} is a random process whose rea\-li\-za\-tion consists of a list of discrete events localized in time, $\cbr{t_i}$ with $t_i \in \RR^+$. Equivalently, a given temporal point process can be re\-pre\-sen\-ted as a counting process, $N(t)$, which records the number of events before time $t$. An important way to characterize temporal point processes is via the conditional intensity function $\lambda(t)$, a stochastic model for the time of the next event given all the previous events. Formally, $\lambda(t)\rd t$ is the conditional probability of observing an event in a small window $[t, t + \rd t)$ given the history $\Hcal(t)$ up to $t$ and that the event has not happen before $t$, \ie,

$\lambda(t)\rd t := \PP\cbr{\text{event in $[t, t+\rd t)$}|\Hcal(t)} = \EE[\rd N(t) | \Hcal(t)]$ \\
where one typically assumes that only one event can happen in a small window of size $\rd t$, \ie,~$\rd N(t) \in \cbr{0,1}$. Then, given a time $t \geqslant 0$, we can also characterize the conditional probability that no event happens during $[0, t)$ as:
$	S(t) = \exp\Big(-{\scriptsize \int_0^{t}} \lambda(\tau) \, \rd\tau \Big)
$ and the conditional density $p(t)$ that an event occurs at time $t$ is defined as 
\begin{equation}
p(t) = \lambda(t)\exp\Big(-{\scriptsize \int_0^{t}} \lambda(\tau) \, \rd\tau \Big)	
\label{eq:f}
\end{equation}
The function form of the intensity $\lambda(t)$ is often designed to capture the phenomena of interests. We present some representative examples of typical point processes where the intensity has a particularly specified parametric forms.

For example, {\bf Poisson processes} has a constant intensity $\lambda(t) = \mu$. {\bf Hawkes processes}~\citep{Hawkes71} models the mutual excitation between events,~\ie, 
$ \lambda(t) = \mu + \alpha \sum\nolimits_{t_i \in \Hcal(t)} \kappa_{\omega}(t-t_i),$
where $\kappa_{\omega}(t):= \exp(-\omega t)$ is an exponential triggering kernel, $\mu\geqslant 0$ is a baseline intensity. Here, the occurrence of each historical event increases the intensity by a certain amount determined by the kernel $\kappa_{\omega}$ and the weight $\alpha \geqslant 0$, making the intensity history dependent and a stochastic process by itself. {\bf Rayleigh process}~\citep{AalBorGje08} models an increased tendency over time 
$$
\vspace{-1mm}
\lambda(t) = \alpha t,
$$
where $\alpha > 0$ is the weight parameter.

\vspace{-1.5mm}
\section{Deep Coevolutionary Feature Embedding Processes}\label{sec:model}

We present Deep Coevolutionary Network (DeepCoevolve): a generative model for modeling the interaction dynamics between users and items. The high level idea of our model is that we first use RNN to explicitly capture the coevolving nature of users' and items' latent features. Then, based on the compatibility between the users' and items' latent feature, we model the user-item interactions by a multi-dimensional temporal point process. We further parametrize the intensity function by the compatibility between users' and items' latent features.

Given $m$ users and $n$ items, we denote the ordered list of $N$ observed events as $\mathcal{O} = \{e_j = (u_j, i_j, t_j, \qb_j)\}_{j=1}^N$ on time window $[0,T]$, where $u_j \in \{1, \ldots, m\}$, $i_j \in \{1, \ldots, n\}$, $t_j \in \RR^+$, $0\leqslant t_1\leqslant t_2 \ldots\leqslant T$. This represents the interaction between user $u_j$, item $i_j$ at time $t_j$, with the interaction context $\qb_j \in \RR^d$. Here $\qb_j$ can be a  high dimension vector such as the text review, or simply the embedding of static user/item features such as user's profile and item's categorical features. For notation simplicity, we define $\Ocal^u = \{e_j^u = (i_j^u, t_j^u, \qb_j^{u})\}$ as the ordered listed of all events related to user $u$, and $\Ocal^i = \{e_j^i = (u^i_j, t^i_j, \qb^i_j)\}$ as the ordered list of all events related to item $i$. We also set $t_0^i = t_0^u = 0$ for all the users and items.  

\vspace{-2mm}
\subsection{Deep coevolutionary network}
As the standard setting in recommendation systems, for each user $u$ and item $i$, we use the low-dimension vector $\fb_u(t)\in\RR^k$ and $\gb_{i}(t)\in\RR^k$ to represent their latent feature embedding at time $t$,~\ie, user's interest and item's property.

These embedding vectors change over time. Specifically, we model the evolution of embeddings $\fb_u(t)$ and $\gb_i(t)$ using two update functions. These embeddings are initialized as 0, and then the updates are carried out whenever a user interacts with an item. The amount of the updates are determined by neural networks which aggregates historical information of user and item embeddings, and interaction time and context. The specific form of the two parallel feature embedding updates are described below. 

\hspace{-6mm}
\fbox{
 \parbox{1.0\linewidth}{
 {\bf DeepCoevolve:} \\
 {\bf Users' embedding update.} For each user $u$,  we formulate the corresponding embedding $\fb_u(t)$ after user $u$'s $k$-th event $e_k^u = (i_k^u, t_k^u, \qb_k^u)$ as follows
 \begin{align}
 \fb_u(t_k^u) & = \sigma\bigg( \underbrace{\Wb_1 (t_k^u- t_{k-1}^u)}_{\text{temporal drift}} + \underbrace{\Wb_2 \fb_u(t_{k-1}^u)}_{\text{self evolution}} + \nonumber \\ & \underbrace{\Wb_3 \gb_{i_k}(t_{k}^u-)}_{\text{co-evolution: item feature}} + \underbrace{\Wb_4 \qb^{u}_k}_{\text{interaction feature}} \bigg). 
 \label{eq:u} 
 \end{align}
 {\bf Items' embedding update.} For each item $i$, model $\gb_i(t)$ after item $i$'s  $k$-th event $e_k^i = (u_k^i, t_k^i, \qb_k^i)$  as follows:
 \begin{align}
 \gb_i(t_k^i) & = \sigma\bigg( \underbrace{\Vb_1 (t_k^i- t_{k-1}^i)}_{\text{temporal drift}} + \underbrace{\Vb_2 \gb_i(t_{k-1}^i)}_{\text{self evolution}} + \nonumber \\ & \underbrace{\Vb_3 \fb_{u_k}(t_{k}^i-)}_{\text{co-evolution: item feature}}  + \underbrace{\Vb_4 \qb^{i}_k}_{\text{interaction feature}} \bigg)\label{eq:i}, 
 \end{align}
 }  
}
where $t-$ means the time point just before time $t$, $\Wb_4,\Vb_4 \in \Rb^{k \times d}$ are the embedding matrices mapping from the explicit high-dimensional feature space into the low-rank latent feature space and $\Wb_1,\Vb_1 \in \RR^k$, $\Wb_2,\Vb_2,\Wb_3,\Vb_3 \in \RR^{k\times k}$ are weights parameters. $\sigma(\cdot)$ is activation function, such as commonly used Tanh or Sigmoid function. For simplicity, we use basic recurrent neural network to formulate the recurrence structure, but it is also straightforward to design more sophisticated structured using GRU or LSTM to gain more expressive power.
Figure~\ref{fig:demo} summarizes the key update equations of our model given a new event.  
Each update incorporates four terms, namely terms related to temporal drift, self evolution, coevolution and interaction features. 

The rationale for designing these terms is explained below: 
\begin{itemize}[leftmargin=*,nosep,nolistsep]
\item {\bf Temporal drift}. The first term is defined based on the time difference between consecutive events of specific user or item. It allows the  users' basic feature (\eg, personalities) and items' basic property (\eg, price, description) to smoothly change over time. Such changes of basic features normally are caused by external influences.

\item {\bf Self evolution}. The current user feature should also be influenced by its feature at the earlier time. This captures the intrinsic evolution of user/item features. For example, a user's current interest should be related to his/her interest two days ago. 
  
\item {\bf User-item coevolution}. This term captures the phenomenon that user and item embeddings are mutually dependent on each other. First, a user's embedding is determined by the latent features of the items he interacted with. At each event time $t_k$, the item embedding influences the update of the user embedding. Conversely, an item's embedding is determined by the feature embedding of the user who just interacts with the item. 

\item {\bf Interaction features}. The interaction feature is the additional information happened in the user-item interactions. For example, in online discussion forums such as Reddit, the interaction features are the posts and comments. In online review sites such as Yelp, the features are the reviews of the businesses. This term models influence of interaction context on user and item embeddings. For example, the contents of a user's post to a Reddit science group are really original and interesting, which sets the future direction of the group. 

\end{itemize}

\vspace{-3mm}
\subsection{Understanding coevolutionary embeddings}

\begin{figure*}[t!]
\small
\centering
\begin{tabular}{cc}
\begin{subfigure}[t]{0.45\textwidth}
\centering
  \includegraphics[scale=0.28]{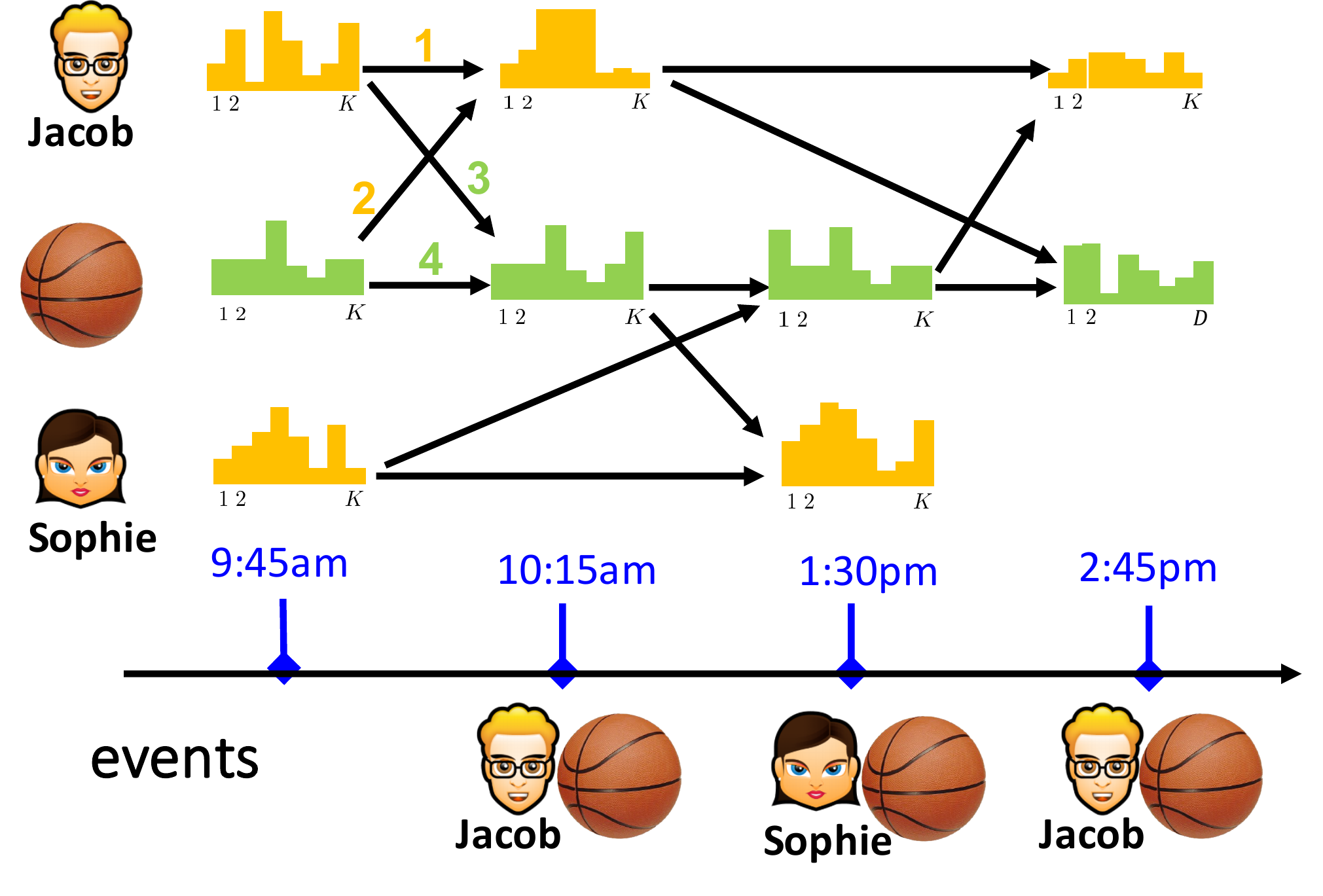}
  \caption{Graph of embedding computation \label{fig:comp_embed}}
\end{subfigure}
&
\begin{subfigure}[t]{0.45\textwidth}
\centering
  \includegraphics[scale=0.28]{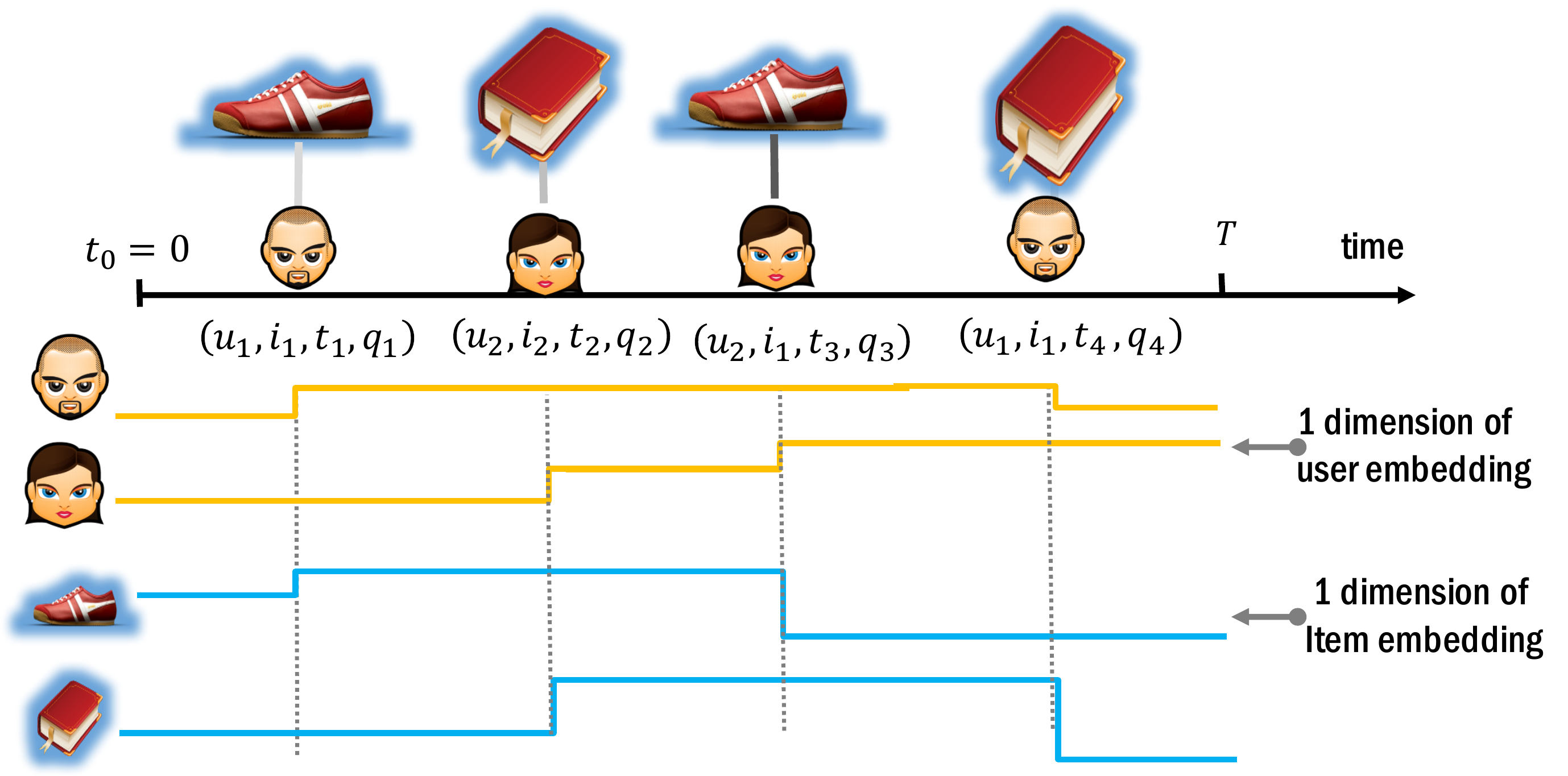}
  \caption{Dependency between events \label{fig:integral1}}
\end{subfigure}
\end{tabular}
\vspace{-4mm}
\caption{(a) The arrows indicate the dependency structures in the embedding updates,~\eg, Jacob interacts with basketball at 10:15am. Then the feature embeddings are updated: the new feature embedding at 10:15 am is influenced by his previous feature embedding and the basketball's previous feature embedding at 9:45am (arrow 1 and 2); the basketball's feature embedding is also influenced by Jacob's previous feature embedding and its previous embedding feature (arrow 3 and 4). (b) A user or item's feature embedding is piecewise constant over time and will change \emph{only} after an interaction event happens. Only one dimension of the feature embedding is shown.}
\label{fig:intensity}
\vspace{-4mm}
\end{figure*}

Although the recurrent updates in~\eq{eq:u} and~\eq{eq:i} involve only the user and item pairs directly participating in that specific interaction, the influence of a particular user or item can propagate very far into the entire bipartite network. Figure~\ref{fig:comp_embed} provides an illustration of such cascading effects. It can be seen that a user's feature embedding can influence the item feature embedding he directly interacts with, then modified item feature embedding can influence a different user who purchases that item in a future interaction event, and so on and so forth through the entire network. 

Since the feature embedding updates are event driven, both the user and item's feature embedding processes are piecewise constant functions of time. These embeddings are \emph{changed only if an interaction event happens}. That is a user's attribute changes only when he has a new interaction with some item. This is reasonable since a user's taste for music changes only when he listens to some new or old musics. Similarly, an item's attribute changes only when some user interacts with it. Such piecewise constant feature embeddings are illustrated in Figure~\ref{fig:integral1}. 

Next we show how to use the feature embeddings to model the complex user-item interaction event dynamics.

\vspace{-3mm}
\subsection{Intensity function as the compatibility between embeddings}
We model the repeated occurrences of all users interaction with all items as a multi-dimensional temporal point process, with each user-item pair as one dimension. Mathematically, we model the intensity function in the $(u,i)$-th dimension (user $u$ and item $i$) as a Rayleigh process:
\begin{equation}
	\lambda^{u, i}(t|t') = \underbrace{\exp \left(\fb_u(t')^\top \gb_i(t')\right)}_{\text{user-item compatibility}} \; \cdot \underbrace{(t - t')}_{\text{time lapse}}
	\label{eq:model}
\end{equation}
where $t > t'$, and $t'$ is the last time point where either user $u$'s embedding or item $i$'s embedding changes before time $t$. 
The rationale behind this formulation is as follows:  
\begin{itemize}[leftmargin=*,nosep,nolistsep]
\item {\bf Time as a random variable.} Instead of discretizing the time into epochs as traditional methods~\citep{ChaRanMciBle15,BhaPhaZhoLee15,GopHofBle15,HidTik15,WanDonNeletal16}, we explicitly model the timing of each interaction event as a random variable, which naturally captures the heterogeneity of the temporal interactions between users and items.
	 
\item {\bf Short term preference.}  The probability for user $u$ to interact with item $i$ depends on the compatibility of their instantaneous embeddings, which is evaluated through the inner product at the last event time $t'$.  Because $\fb_u(t)$ and $\gb_i(t)$ co-evolve through time, their inner-product measures a general representation of the cumulative influence from the past interactions to the occurrence of the current event. The $\exp(\cdot)$ function ensures the intensity is positive and well defined.
	 
\item {\bf Rayleigh time distribution.} The user and item  embeddings are \emph{piecewise constant}, and we use the time lapse term to make the intensity \emph{piecewise linear}. This form leads to a Rayleigh distribution for the time intervals between consecutive events in each dimension. It is well-adapted to modeling fads~\cite{AalBorGje08}, where the likelihood of generating an event  rises to a peak and then drops extremely rapidly. Furthermore, it is computationally easy to obtain an analytic form of this likelihood. One can then use it to make item recommendation by finding the dimension that the likelihood function reaches the peak.
\end{itemize}

\begin{figure*}[t]
\small
\centering
\begin{subfigure}[t]{0.3\textwidth}
\centering
  \includegraphics[scale=0.25]{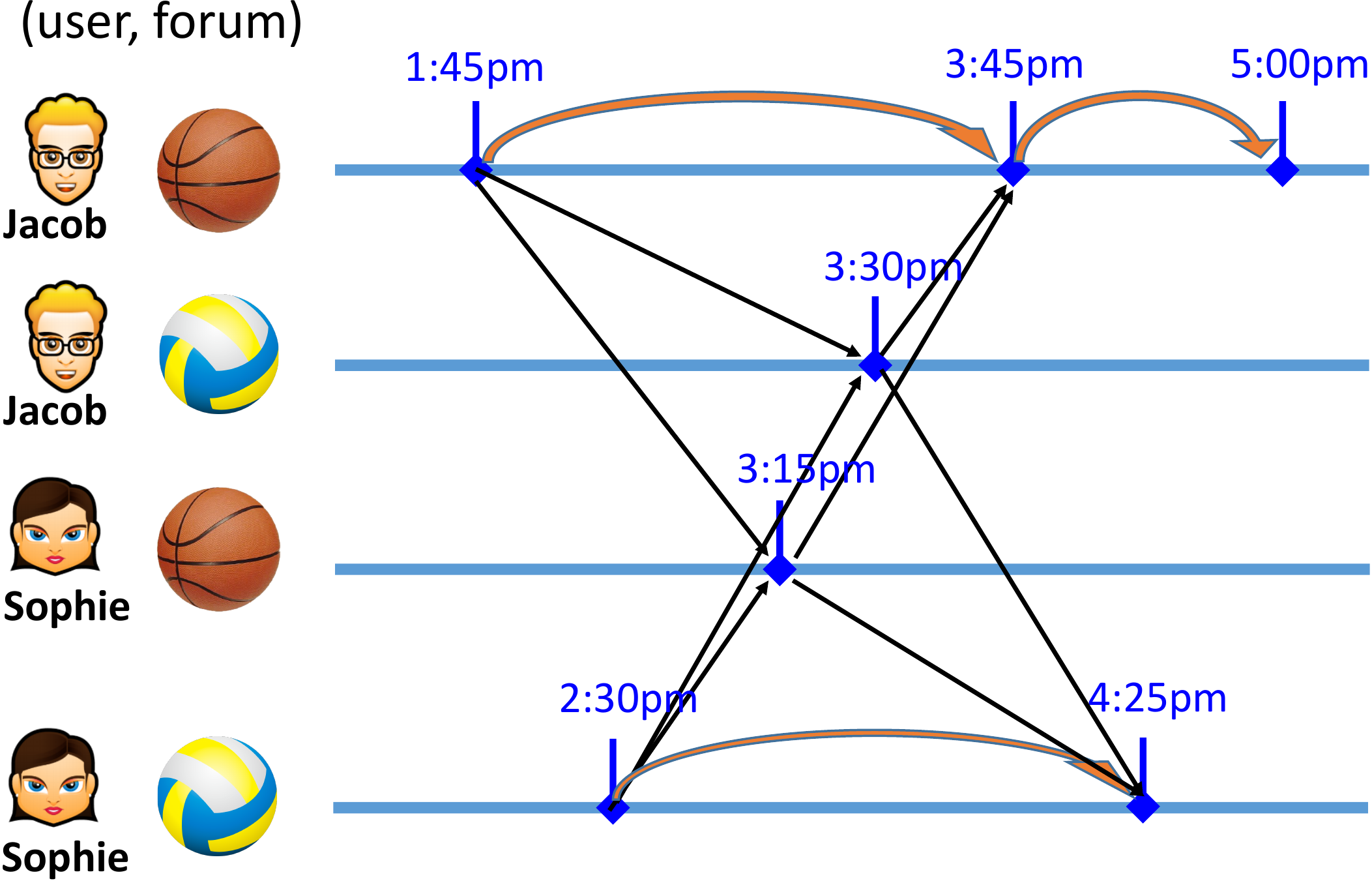}
  \caption{Dependency between events \label{fig:int_dep}}
\end{subfigure}
\begin{subfigure}[t]{0.3\textwidth}
\centering
  \includegraphics[scale=0.25]{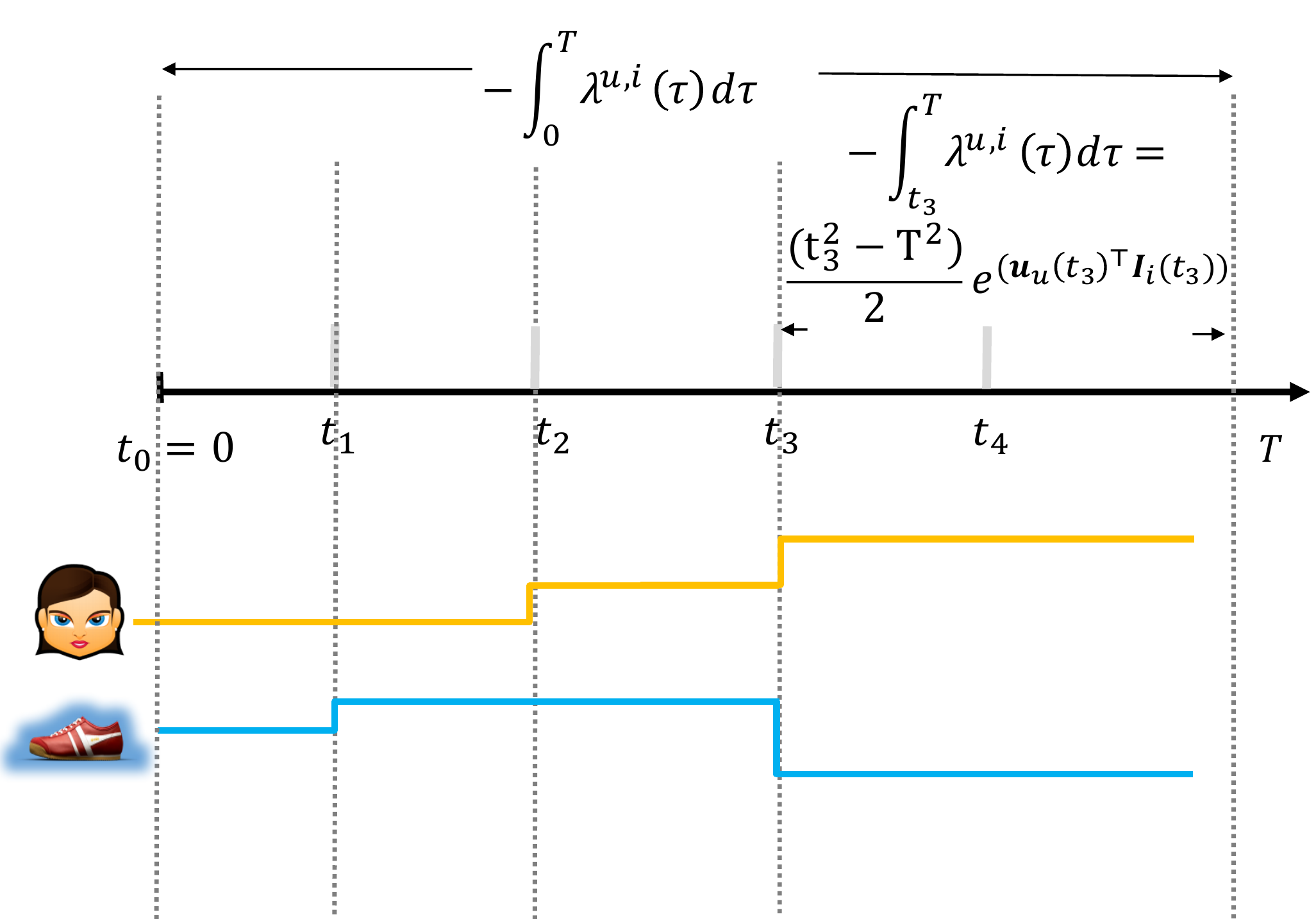}
  \caption{Survival probability computation} \label{fig:survival}
\end{subfigure}
\begin{subfigure}[t]{0.28\textwidth}
  \includegraphics[width=\textwidth, trim=0 30 0 0]{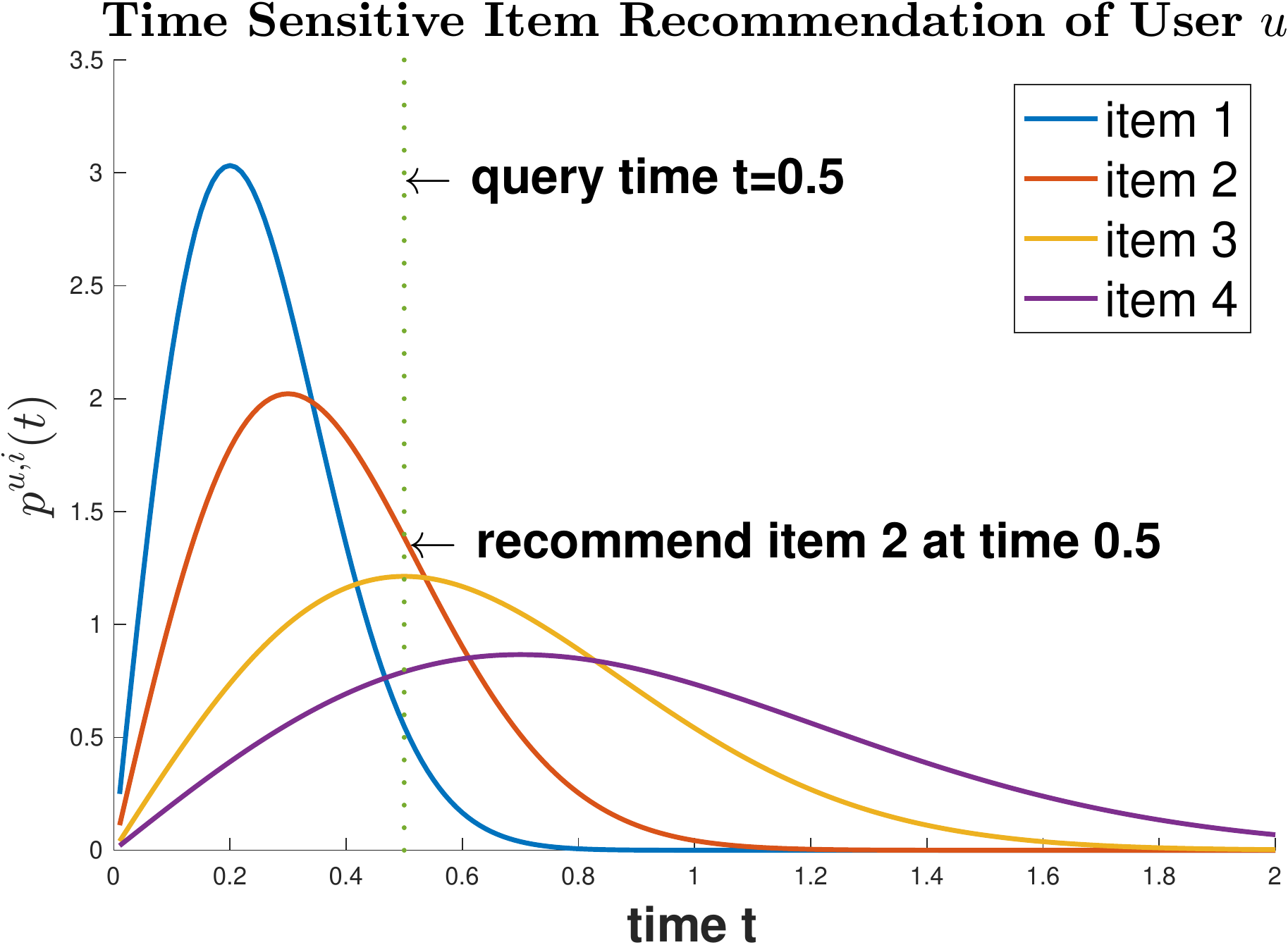}
  \caption{Item recommendation}
  \label{fig:item_pred}
\end{subfigure}
\vspace{-2mm}
\caption{(a) The events dependency for two users and two forums (items). It shows how event at one dimension influence other dimensions. Each orange arrow represents the dependency within each dimension, and the black arrow denotes the cross-dimension dependency. For example, Sophie interacts with volleyball at 2:30pm, and this event changes the volleyball embedding, thus will affect Jacob's visit at 3:30pm. (b) Survival probability for a user-item pair $(u, i)$. The integral $\int_0^T\lambda^{u,i}(\tau|\tau')\rd \tau$ is decomposed into 4 inter-event intervals separated by $\cbr{t_0,\cdots,t_3}$. (c) Illustration of item recommendation. }
\label{fig:intensity}
\vspace{-2mm}
\end{figure*}

\vspace{-2mm}
\section{Efficient Learning for Deep Coevolutionary Network}
\vspace{-0.5mm}
\label{sec:learning}
In this section, we will first introduce the objective function, and then propose an efficient learning algorithm.

\vspace{-2mm}
\subsection{Objective  function}
\vspace{-1mm}

With the parameterized intensity function in \eq{eq:model}, we can sample events according to it. Due to the interdependency between the feature embeddings and the propagation of influence over the interaction network, the different dimensions of the point process can intricate dependency structure. Such dependency allows sophisticated feature diffusion process to be modeled. Figure~\ref{fig:int_dep} illustrates the dependency structures of the generated events.  

Given a sequence of events observed in real world, we can further estimate the parameters of the model by maximizing the likelihood of these observed events. Given a set of $N$ events, the joint negative log-likelihood can be written as~\citep{DalVer2007}:
\begin{equation}
\vspace{-1mm}
\ell = \underbrace{- \sum_{j=1}^N \log\rbr{\lambda^{u_j,i_j}(t_j|t'_j)}}_{\text{happened events}}
	+ \underbrace{\sum_{u=1}^m\sum_{i=1}^n \int_0^T \lambda^{u,i}(\tau|\tau')\, \rd \tau}_{\text{survival of not-happened events}}\label{eq:data_loglikelihood} 
\end{equation}
We can interpret it as follows: (i) the negative {intensity} summation term ensures the probability of all interaction events is maximized; (ii) the second survival probability term penalizes the \emph{non-presence} of an interaction between all possible user-item pairs on the observation window. Hence, our framework not only explains why an event happens, but also why an event did not happen.

Due to the co-evolution nature of our model, it is a very challenging task to learn the model parameters since the bipartite interaction network is time-varying. Next, we will design an efficient learning algorithm for our model.

\vspace{-3mm}
\subsection{Efficient learning algorithm}
\vspace{-1mm}
We propose an efficient algorithm to learn the parameters  $\cbr{\Vb_i}_{i=1}^4$ and $\cbr{\Wb_i}_{i=1}^4$. 
The Back Propagation Through Time (BPTT) is the standard way to train a RNN. To make the back propagation tractable, one typically needs to do truncation during training. However, due to the novel co-evolutionary nature of our model, all the events are related to each other by the user-item bipartite graph (Figure~\ref{fig:int_dep}), which makes it hard to decompose. 

Hence, in sharp contrast to works~\citep{HidKarBalTik16, DuDaiTriUpaGomSon16} in sequential data where one can easily break the sequences into multiple segments to make the BPTT trackable, it is a \emph{challenging} task to design BPTT in our case. To efficiently solve this problem, we first order all the events globally and then do mini-batch training in a sliding window fashion. Each time when conducting feed forward and back propagation, we take the consecutive events within current sliding window to build the computational graph. Thus in our case the truncation is on the global timeline, compared with the truction on individual independent sequences in prior works. 

Next, we explain our procedure in detail. Given a mini-batch of $M$ ordered events $\tilde{\Ocal} = \{e_j\}_{j=1}^M$, we set the time span to be $[T_0 = t_1, T = t_M]$. Below we show how to compute the intensity and survival probability term in the objective function~(\ref{eq:data_loglikelihood}) respectively.
\begin{itemize}[leftmargin=*,nosep,nolistsep]
  \item {\bf Computing the intensity function.} Each time when a new event $e_j$ happens between $u_j$ and $i_j$, their corresponding feature embeddings will evolve according to a computational graph, illustrated in Figure~\ref{fig:comp_embed}. Due to the change of feature embedding, all the dimensions related to $u_j$ or $i_j$ are also influenced and the intensity functions for these dimensions change consequently. Such cross-dimension  dependency of influence is further shown in Figure~\ref{fig:int_dep}. In our implementation, we first compute the corresponding intensity $\lambda^{u_j,i_j}(t_j|t_j')$ according to (\ref{eq:model}), and then update the embedding of $u_j$ and $i_j$. This operation takes $O(M)$ complexity, and is independent to the number of users or items. 

  \item {\bf Computing the survival function.} To compute the survival probability $-\int_{T_0}^T \lambda^{u, i}(\tau|\tau')\rd \tau$ for each pair $(u, i)$, we first collect all the time stamps $\{t_k\}$ that have events related to either $u$ or $i$. For notation simplicity, let $|\{t_k\}| = n_{u, i}$ and $t_1 = T_0, t_{n_{u, i}} = T$. Since the embeddings are piecewise constant, the corresponding intensity function is piecewise linear according to~(\ref{eq:model}). Thus, the integration is decomposed into each time interval where the intensity is linear,~\ie,
  \begin{align}
  \vspace{-2mm}
          \int_{T0}^T \lambda^{u, i}(\tau|\tau') \rd\tau & = \sum_{k=1}^{n_{u, i} - 1} \int_{t_{k}}^{t_{k + 1}} \lambda^{u, i}(\tau|\tau') \rd\tau \\
          & = \sum_{k=1}^{n_{u, i} - 1} (t_{k+1}^2 - t_k^2) \exp\big(\fb_u(t_k)^\top \gb_i(t_k)\big)      
  \end{align}
  Figure~\ref{fig:survival} illustrates the details of computation. 

  Although the survival probability term exists in closed form, it is still expensive to compute it for each user item pair. Moreover, since the user-item interaction bipartite graph is very sparse, it is not necessary to monitor each dimension in the stochastic training setting. To speed up the computation, we propose a novel random-sampling scheme as follows.

  Note that the intensity term in the objective function~\eq{eq:data_loglikelihood} tries to maximize the inner product between user and item that has interaction event, while the survival term penalizes over all other pairs of inner products. We observe that this is similar to Softmax computing for classification problem. Hence, inspired by the noise-contrastive estimation method (NCE)~\citep{GutHyv12} that is widely used in language models~\citep{MniKav13}, we keep the dimensions that have events on them, while randomly sample dimensions without events in current mini-batch to speed up the computation. 
\end{itemize}

Finally, another challenge in training lies in the fact that the user-item interactions vary a lot across mini-batches, hence the corresponding computational graph also changes greatly. To make the training efficient, we use the graph embedding framework~\citep{DaiDaiSon16} which allows training deep learning models where each term in the objective has a different computational graphs but with shared parameters. The Adam Optimizer~\citep{KinBa14} and gradient clip is used in our experiment. 

\begin{table*} [t]
\small
\centering
\caption{Comparison with different methods.}
	\renewcommand{\tabcolsep}{2.0pt}
\vspace{-2mm}
\begin{tabular}{c|c c c c c c c c}
 \hline
Method & \small DeepCoevolve &\small LowRankHawkes &\small Coevolving &\small PoissonTensor  &\small TimeSVD++ &\small  FIP &\small STIC \\ \hline \hline
\small Continuous time & $\surd$ & $\surd$ & $\surd$ & & & & $\surd$\\\hline
\small Predict Item  &$\surd$ & $\surd$ & $\surd$ & $\surd$ & $\surd$ & $\surd$ & \\\hline
\small Predict Time &$\surd$ & $\surd$ & $\surd$& $\surd$& & &$\surd$\\\hline
\small Computation & \small RNN &\small Factorization &\small Factorization &\small Factorization &\small Factorization &\small Factorization &\small HMM\\\hline
\end{tabular}
\vspace{-2mm}
\label{tbl:baseline}
\end{table*}

\vspace{-2mm}
\section{Prediction with DeepCoevolve} 

Since we use DeepCoevolve to model the intensities of multivariate point processes, our model can make two types of predictions, namely next item prediction and event time prediction. The precise event time prediction is especially novel and not possible by most prior work. More specifically, 
\begin{itemize}[leftmargin=*,nosep,nolistsep]
  \item \emph{Next item prediction}. Given a pair of user and time $(u, t)$, we aim at answering the following question: \emph{what is the item the user $u$ will interact at time $t$?} To answer this problem, we rank all items in the descending order in term of the value of the corresponding conditional density at time $t$, 
  \begin{equation}
  \vspace{-1mm}
          p^{u,i}(t) = \lambda^{u,i}(t)S^{u,i}(t)
          \label{eq:item_pred}
  \end{equation} 
  and the best prediction is made to the item with the largest conditional density. Using this formulation, at different point in time, a different prediction/recommendation can be made, allowing the prediction to be time-sensitive. Figure~\ref{fig:item_pred} illustrates such scenario. 
  
  
  \item \emph{Time prediction}. Given a pair of user and item $(u, i)$, we aim at answering the following question: \emph{When this user will interact with this item in the future?} We predict this quantity by computing the expected next event time under $p^{u,i}(t)$. Since the intensity model is a Rayleigh model, the expected event time can be computed in closed form as 
  \begin{equation}
          \EE_{t\sim p^{u,i}(t)}[t] = \sqrt{\frac{\pi}{2\exp\left(\fb_u(t-)^\top \gb_i(t-)\right)}}
          \label{eq:time_pred}
          \vspace{-1mm}
  \end{equation}  
\end{itemize}                    

\vspace{-2mm}
\section{Complexity Analysis}
In this section, we provide an in depth analysis of our approach in terms of the model size, the training and testing complexity. We measure these terms as functions of the number of users, number of items, and the number of events. Other factors, such as dimensionality of latent representations, are treated as constant.  

\begin{itemize}[leftmargin=*,nosep,nolistsep]
  \item {\bf Model Size}. If the baseline profile feature for user and items are not available, we can use one-hot representation of user and items, and the basic feature embedding takes $O($\#user + \#item$)$ parameters. The interaction features (e.g., bag of words features for reviews) are independent of the number of users and number of items. Moreover, the parameters of RNN are also independent of the dataset. Thus, our model is as scalable as the traditional matrix factorization methods. 

  \item {\bf Training Complexity}. Using BPTT with a budget limit, each mini-batch training will only involve consecutive $M$ samples. When an event happens, the embeddings of the corresponding user and item are updated. Thus we need $O(M)$ operations for updating embeddings. For each user item pair, $(n+m)$ dimensions that are correlated with this user-item pair will update their constant intensity functions. Hence, ideally we need $O((n + m)\times M)$ operations to forward the intensity function and survival probability. As mentioned in Section~\ref{sec:learning}, using NCE to sample $C$ dimensions that survive from last event to current event, we can further reduce the computational cost to $C \times M$. Thus in summary, each stochastic training step takes $O(M)$ cost, which is linear to the number of samples in mini-batch. 

  \item {\bf Test/Prediction Complexity}. The item prediction in~\eq{eq:item_pred} requires comparing each item with the current user embedding. Since \eq{eq:item_pred} has a closed form, the complexity is $O(N)$ where $N$ is the number of items. This can further be improved by other methods such as fast inner product search~\cite{BalKolPinSes15}. Since the event time prediction in~\eq{eq:time_pred} is in closed form, the complexity for this prediction task is $O(1)$. 
\end{itemize}

%

\vspace{-3mm}
\section{Experiments}
\begin{figure*}[t]
  \small
  \centering
  \begin{tabular}{cccc}
  \rotatebox{90}{\bf\small~~~{dataset characteristics}}~~
  & \includegraphics[width = 0.32\textwidth]{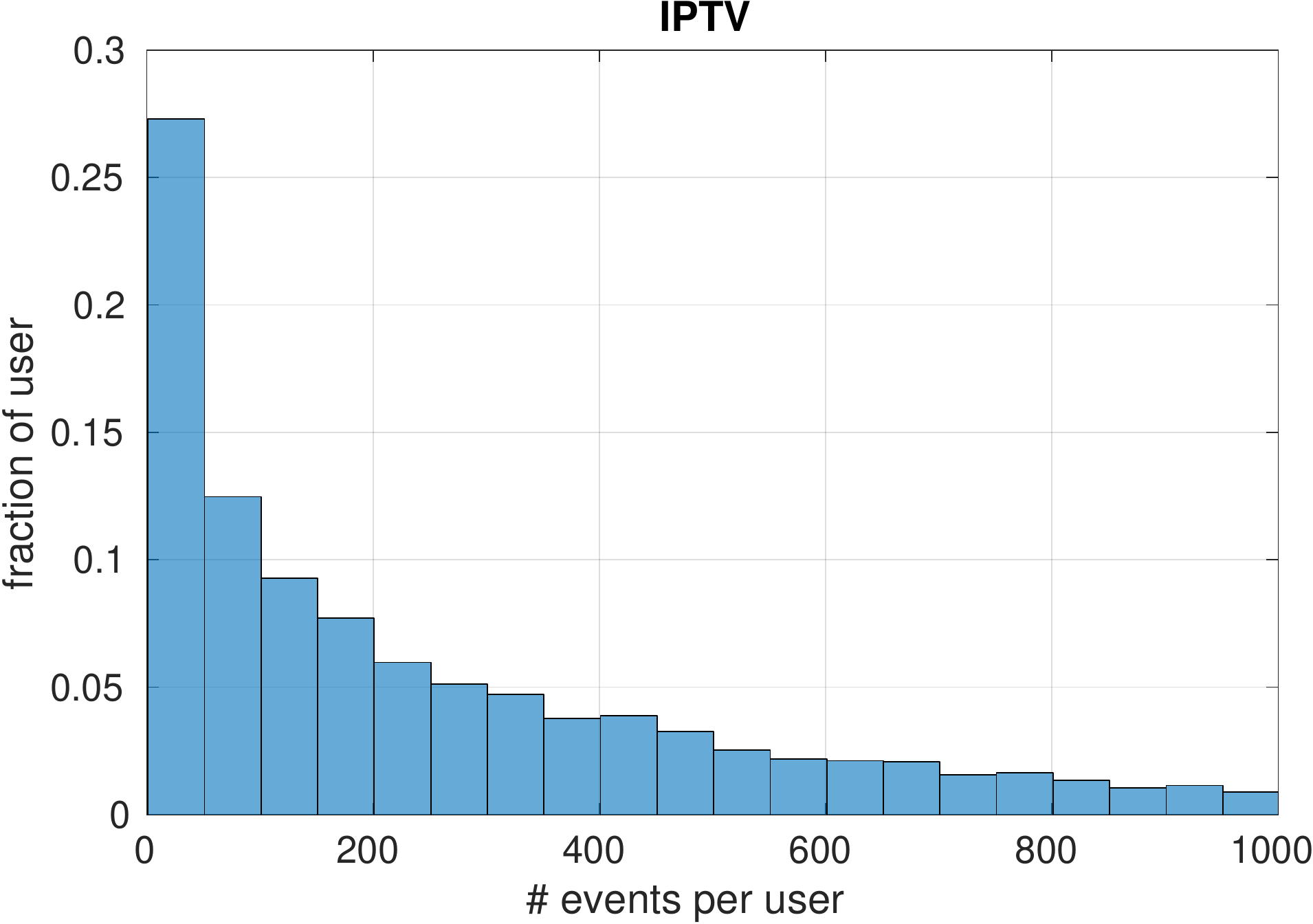}
  \includegraphics[scale=0.26, trim=400 -120 0 0]{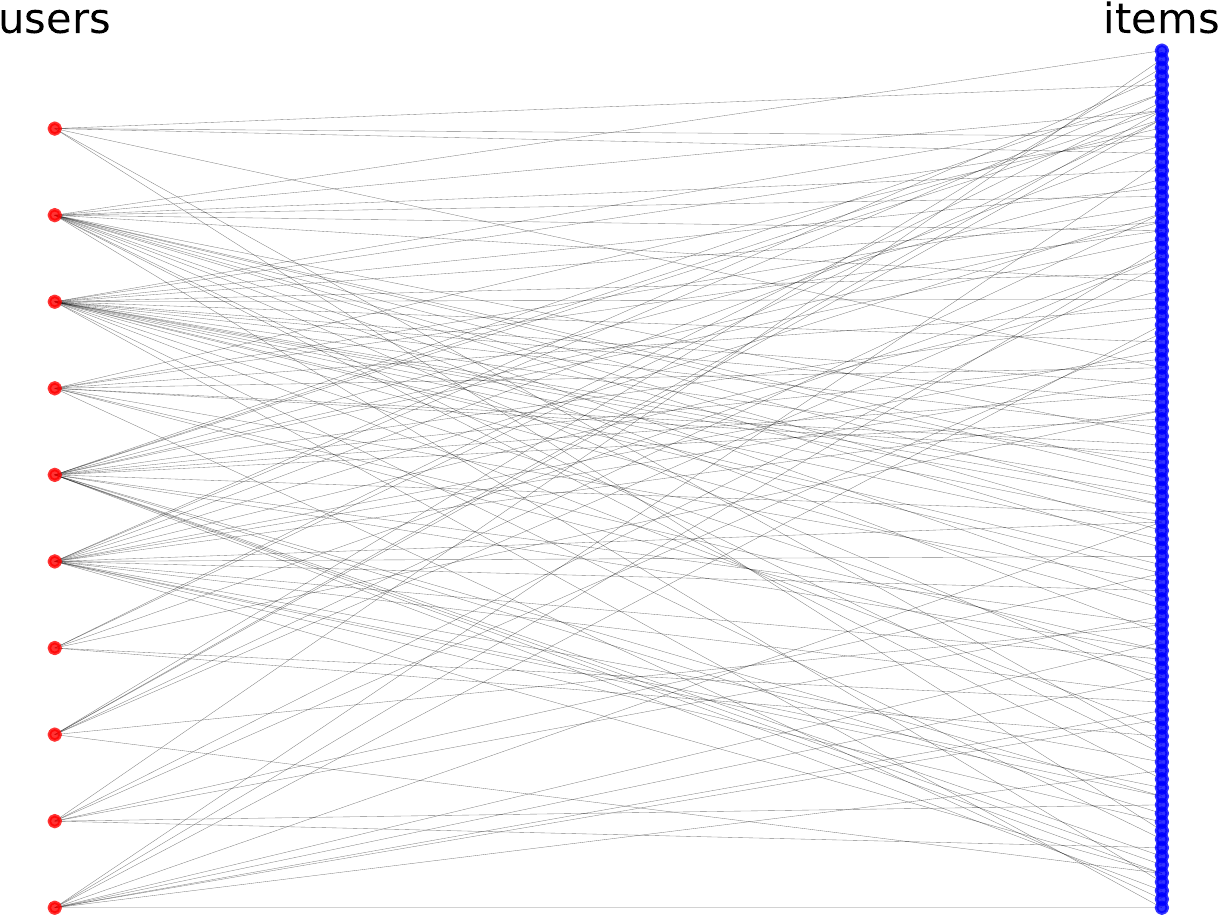}
  & \includegraphics[width = 0.32\textwidth]{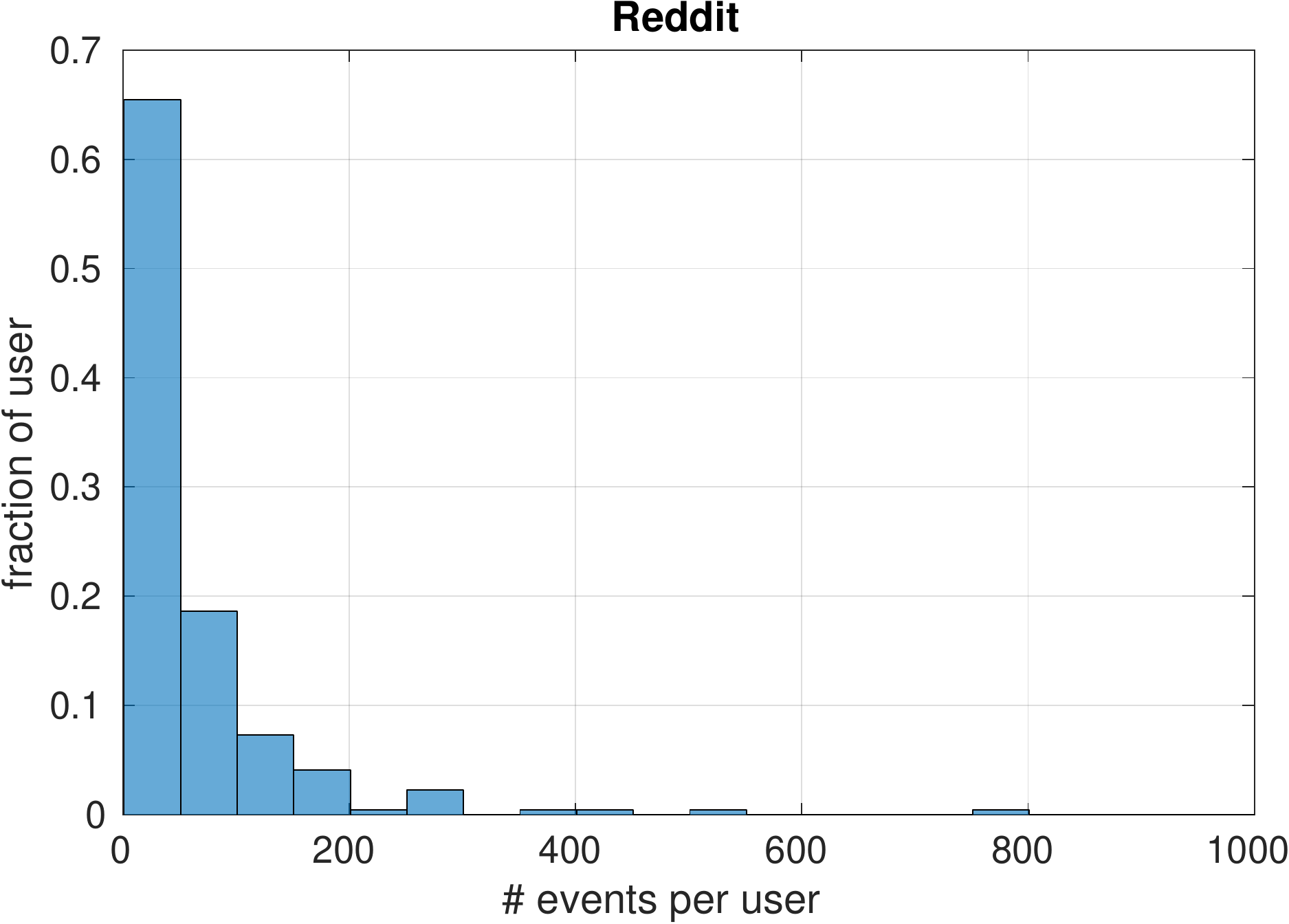}
  \includegraphics[scale=0.26, trim=400 -100 0 0]{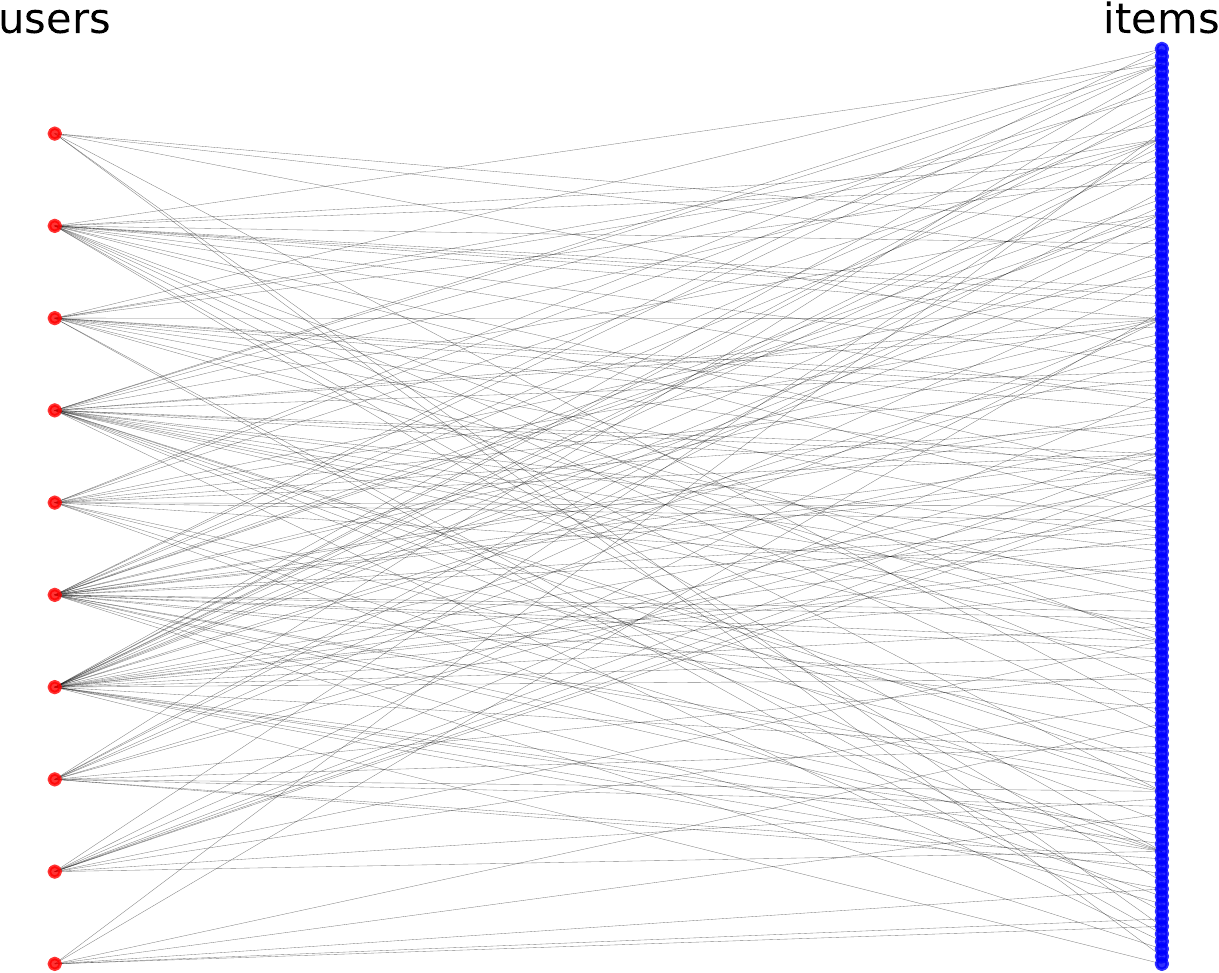}
  & \includegraphics[width = 0.3\textwidth]{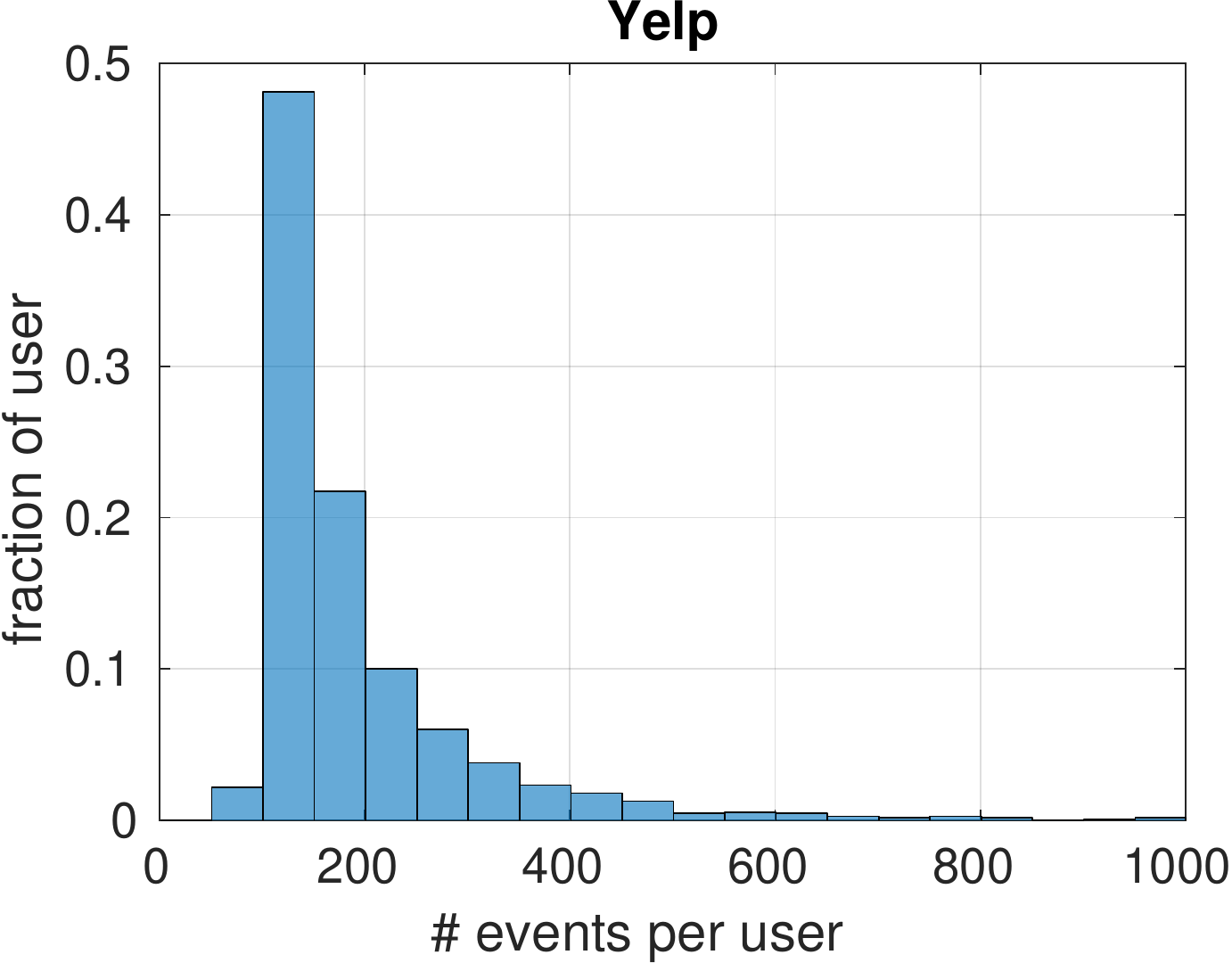}
  \includegraphics[scale=0.26, trim=390 -115 0 0]{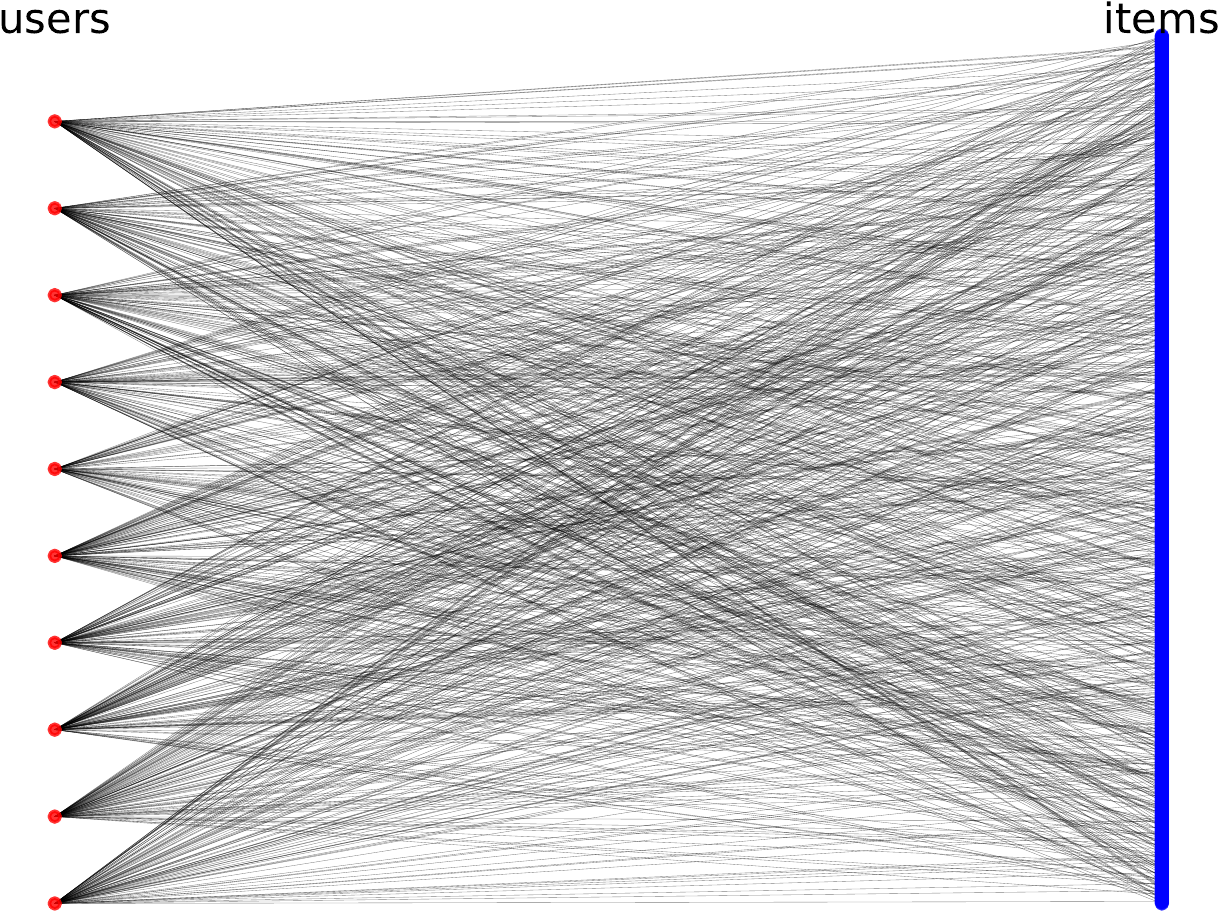}
  \end{tabular}
  \vspace{-4mm}
  \caption{Visualization of the sparsity property in each dataset. The bar plot shows the distribution of number of events per user. Within each plot there is a user-item interaction graph. This graph is generated as follows. For each dataset, we randomly pick 10 users with 100 history events each user and collect all items they have interacted with. The interaction graph itself is a bipartite graph, and we put users on left side, and items on the right side.}
  \vspace{-3mm}
  \label{fig:data_stats}
\end{figure*}

\vspace{-0.5mm}
\label{sec:experiments}
We evaluate our method on three real-world datasets: 
\begin{itemize}[leftmargin=*,nosep,nolistsep]
\item {\bf IPTV.} It contains 7,100 users’ watching history of 436 TV programs in 11 months (Jan 1 - Nov 30 2012), with around 2M events, and 1,420 movie features (including 1,073 actors, 312 directors, 22 genres, 8 countries and 5 years). 
\item {\bf Yelp.} This data was available in Yelp Dataset challenge Round 7. It contains reviews for various businesses from October, 2004 to December, 2015. The dataset we used here contains 1,005 users and 47,924 businesses, with totally 291,716 reviews.
\item {\bf Reddit.} We collected discussion related data on different subreddits (groups) for the month of January 2014. We removed all bot users' and their posts from this dataset. Furthermore, we randomly selected 1,000 users, 1,403 groups, and 14,816 discussions.
\end{itemize}
To study the sparsity of these datasets, we visualize the three datasets in Figure~\ref{fig:data_stats} from two perspectives: (i) the number of events per user, and (ii) the user-item interaction graph.

{\bf Sparsity in terms of the number of events per user.} Typically, the more user history we have, the better results we should expect in the prediction tasks. In IPTV dataset, users have longer length of history than other two datasets. Thus we expect different methods will have the best performance on this dataset. 
 
{\bf Sparsity in terms of diversity of items to recommend.}  If users has similar tastes, the distinct number of items in the union of their history should be small. From the user-item bipartite graph, it is easy to see that Yelp dataset has higher density than the other two datasets, hence higher diversity. The density of the interaction graph also reflects the \emph{variety} of history event for each user. For example, the users in IPTV only have 436 programs to watch, but the users in Yelp dataset can have 47,924 businesses to choose. Also, the Yelp dataset has 9 times more items than IPTV and Reddit dataset in the bipartite graph. This means the users in Yelp dataset has more diverse tastes than users in other two datasets.

Based on the above two facts, we expect that the Yelp dataset is the most challenging one, since it has shorter length of history per user, and much more diversity of the items. 

\vspace{-2.5mm}
\subsection{Competitors}
We compared our \textsc{DeepCoevolve} with the following state-of-arts. Table~\ref{tbl:baseline} summarizes the differences between methods.
\begin{itemize}[leftmargin=*,nosep,nolistsep]
\item {\bf LowRankHawkes}~\citep{DuWanHeetal15}: This is a low rank Hawkes process model which assumes user-item interactions to be independent of each other and does not capture the co-evolution of user and item features.
\item {\bf Coevolving}~\citep{WanDuTriSon16}: This is a multi-dimensional point process model which uses a simple linear embedding to model the co-evolution of user and item features.
\item {\bf PoissonTensor}~\citep{ChiKol12}: Poisson Tensor Factorization has been shown to perform better than factorization methods based on squared loss~\citep{KarAmaBalOli10,XioCheHuaTzu10,WanCheGhoDenetal15} on recommendation tasks. The performance for this baseline is reported using the average of the parameters fitted over all time intervals. 
\item {\bf TimeSVD}++~\citep{Koren09} and {\bf FIP}~\citep{YanLonSmoEtal11}: These two methods are only designed for explicit ratings, the implicit user feedbacks (in the form of a series of interaction events) are converted into the explicit ratings by the respective frequency of interactions with users. 
\item {\bf STIC}~\citep{KapSubKarSriJaiSch15}: it fits a semi-hidden markov model (HMM) to each observed user-item pair and is only designed for time prediction. 
\end{itemize}


\begin{figure*}[t]
\centering
\begin{tabular}{cccc}
\rotatebox{90}{\bf\small~~~~~~~~~~~~~~~~~~~{Item prediction}}~~
& \includegraphics[width = 0.25\textwidth]{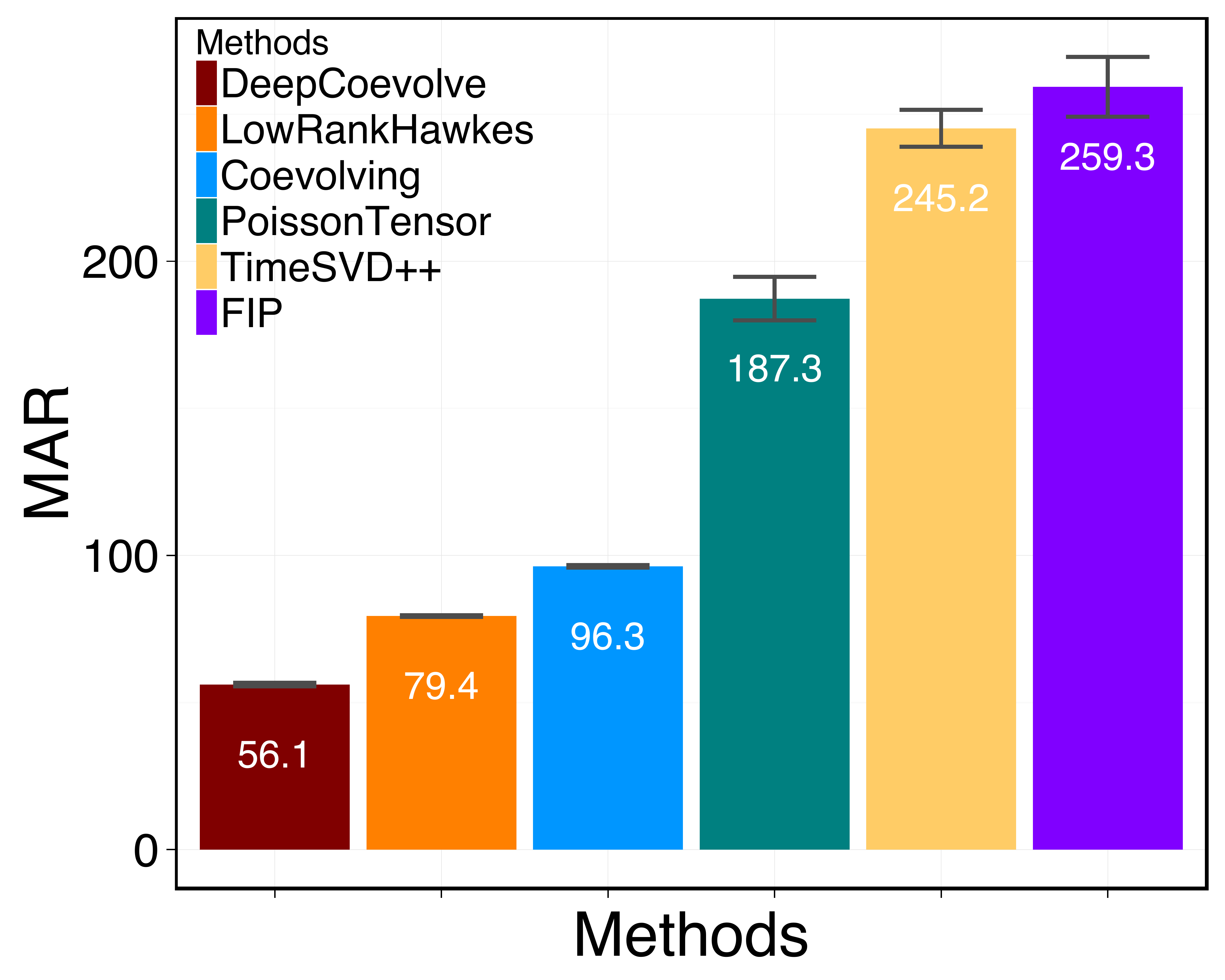}
& \includegraphics[width = 0.25\textwidth]{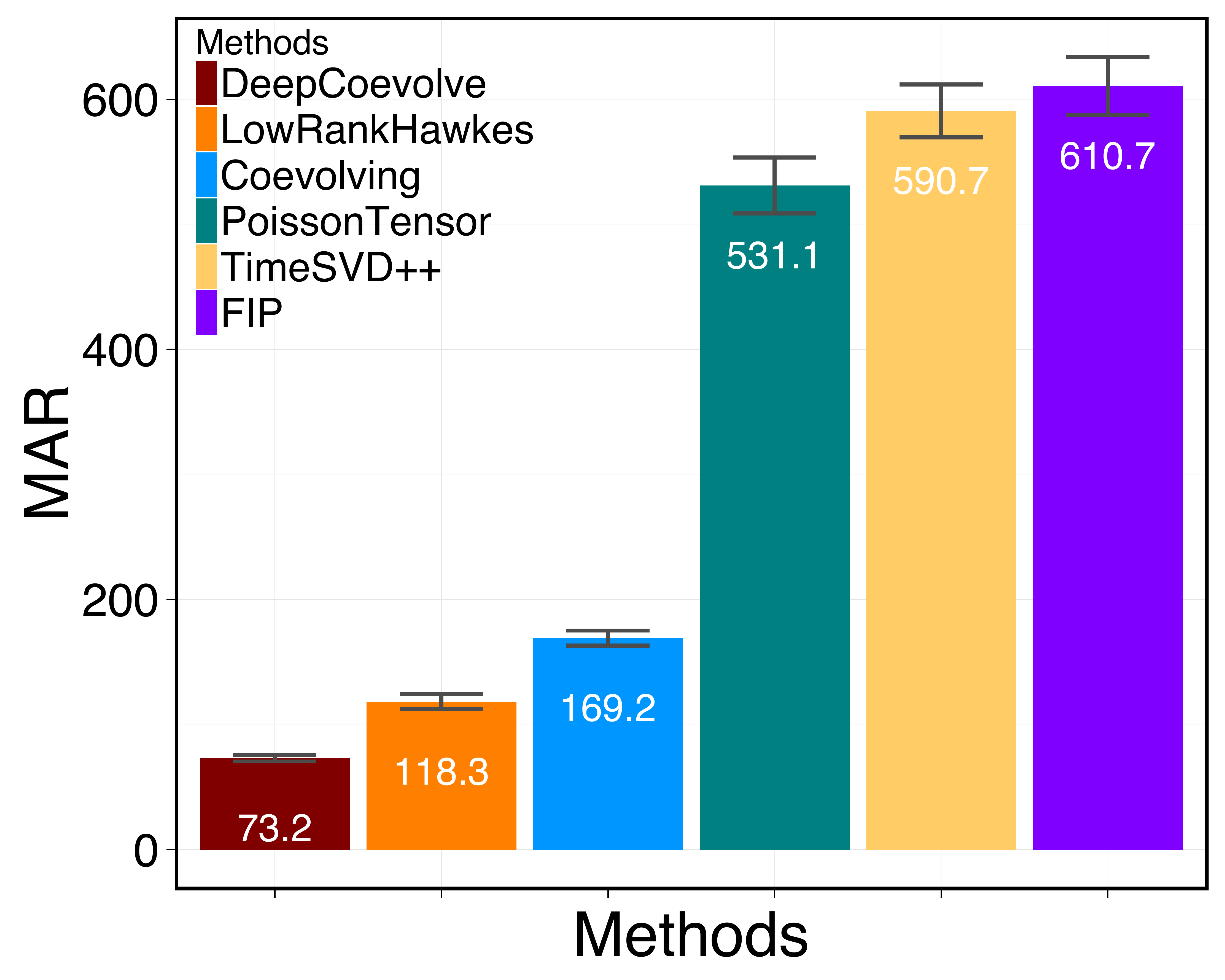}
& \includegraphics[width = 0.25\textwidth]{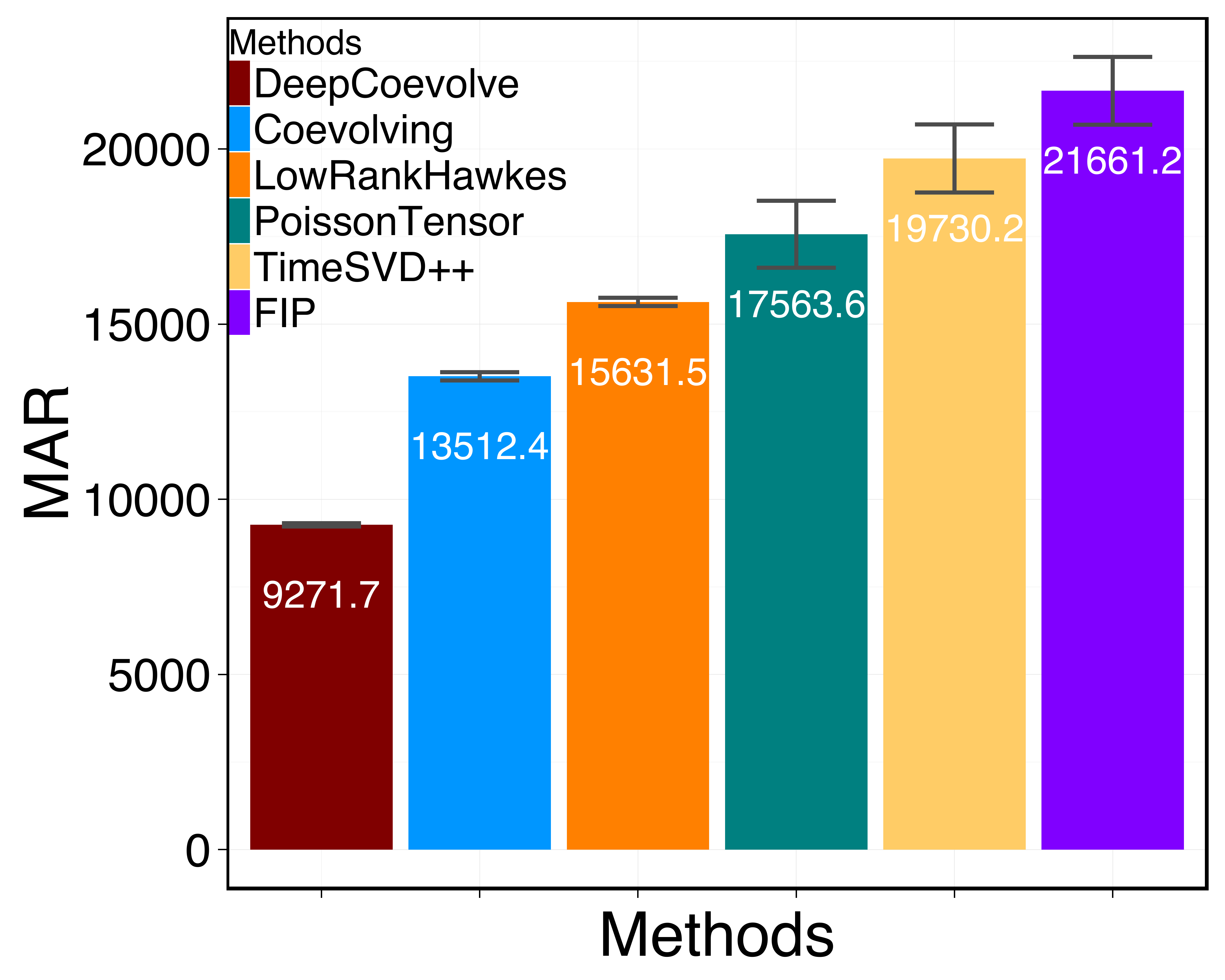}\\
\rotatebox{90}{\bf\small~~~~~~~~~~~~~~~~~{Time prediction}}~~
& \includegraphics[width = 0.25\textwidth]{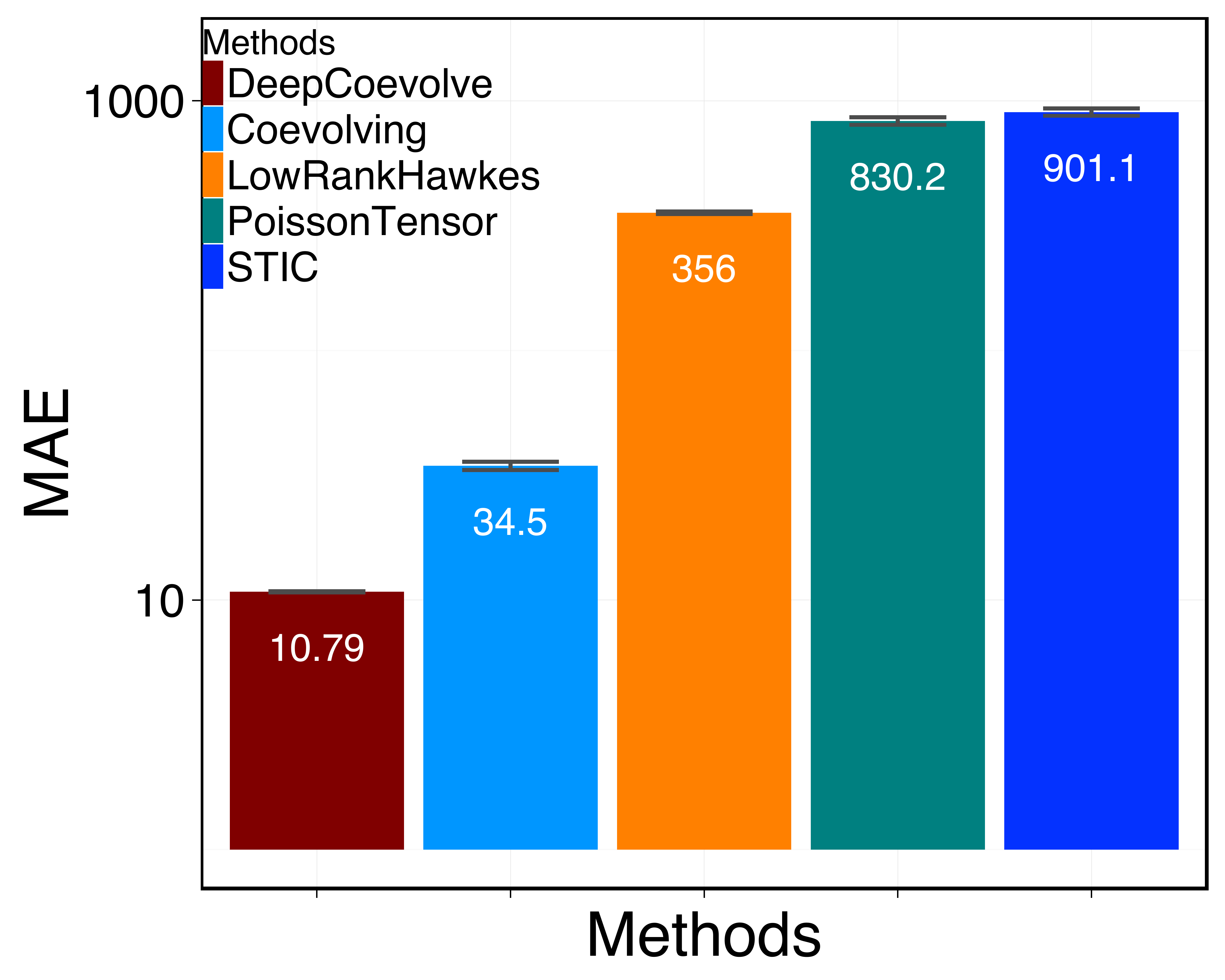}
& \includegraphics[width = 0.25\textwidth]{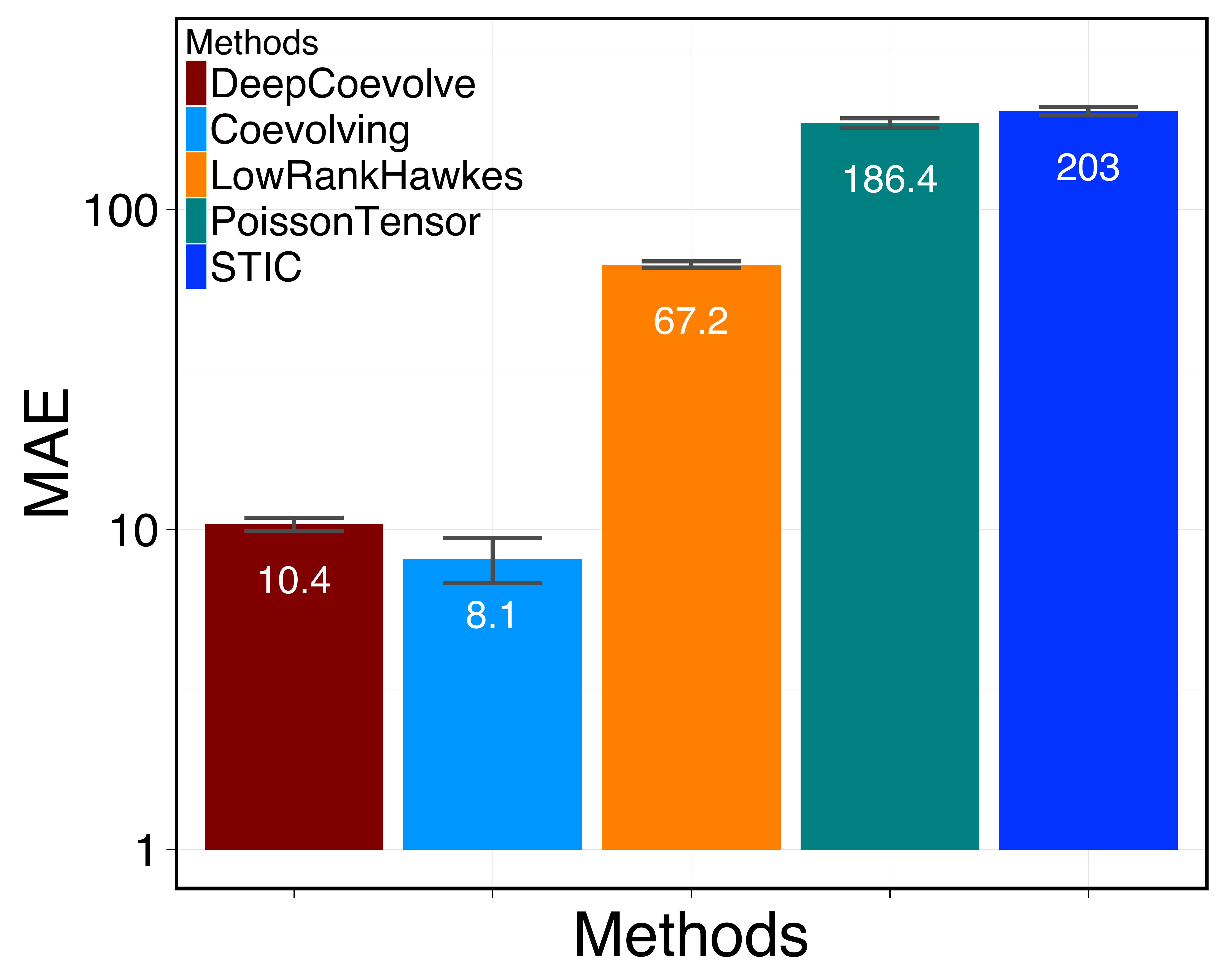}
& \includegraphics[width = 0.25\textwidth]{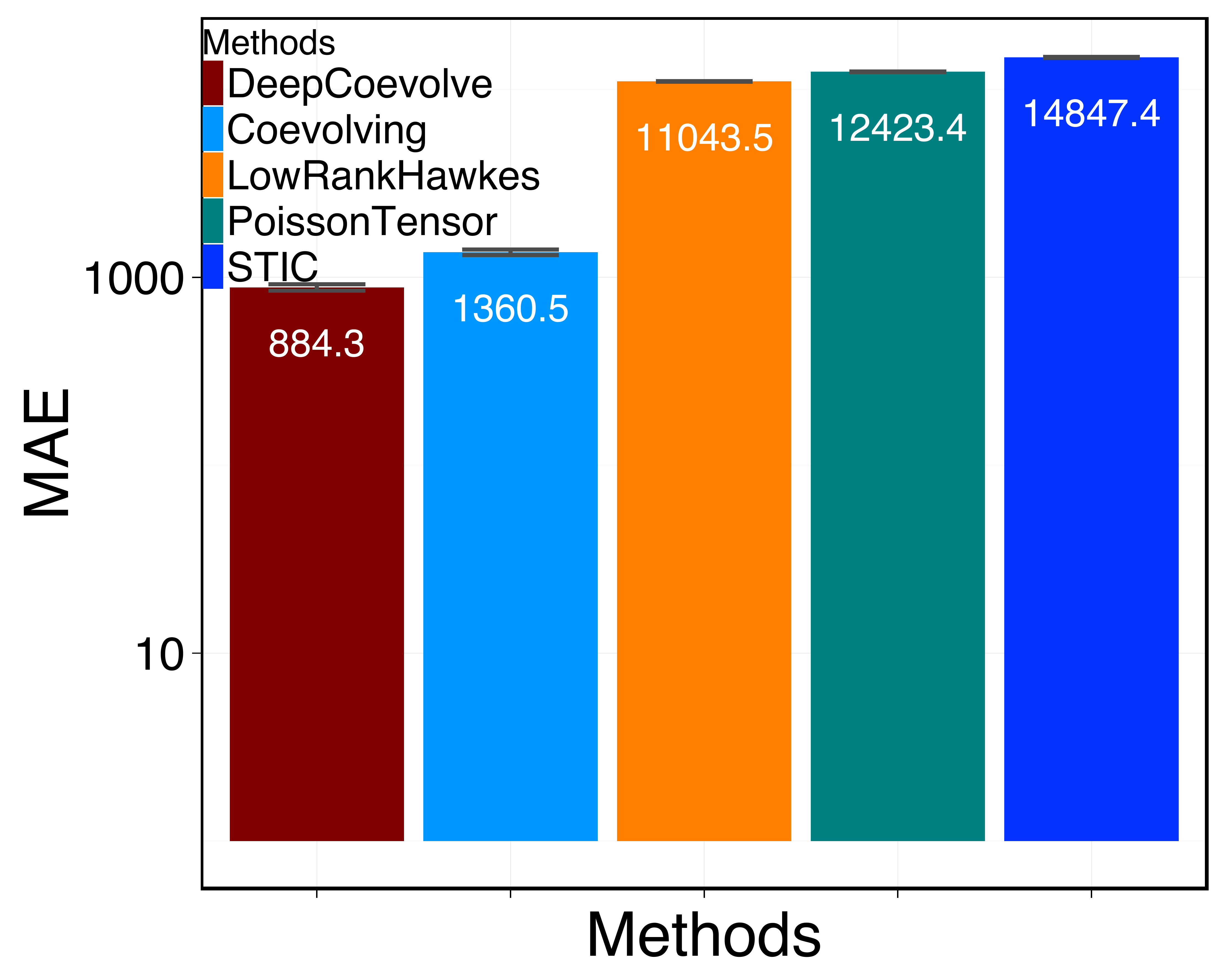}\\
& (a) IPTV	& (b) Reddit & (c) Yelp
\end{tabular}
\vspace{-4mm}
\caption{Prediction results on three real world datasets. The MAE for time prediction is measured in hours.}
\label{fig:exp}
\vspace{-2mm}
\end{figure*}

\begin{figure*}[t]
\small
\centering
\begin{tabular}{ccccc}
& \includegraphics[width = 0.26\textwidth]{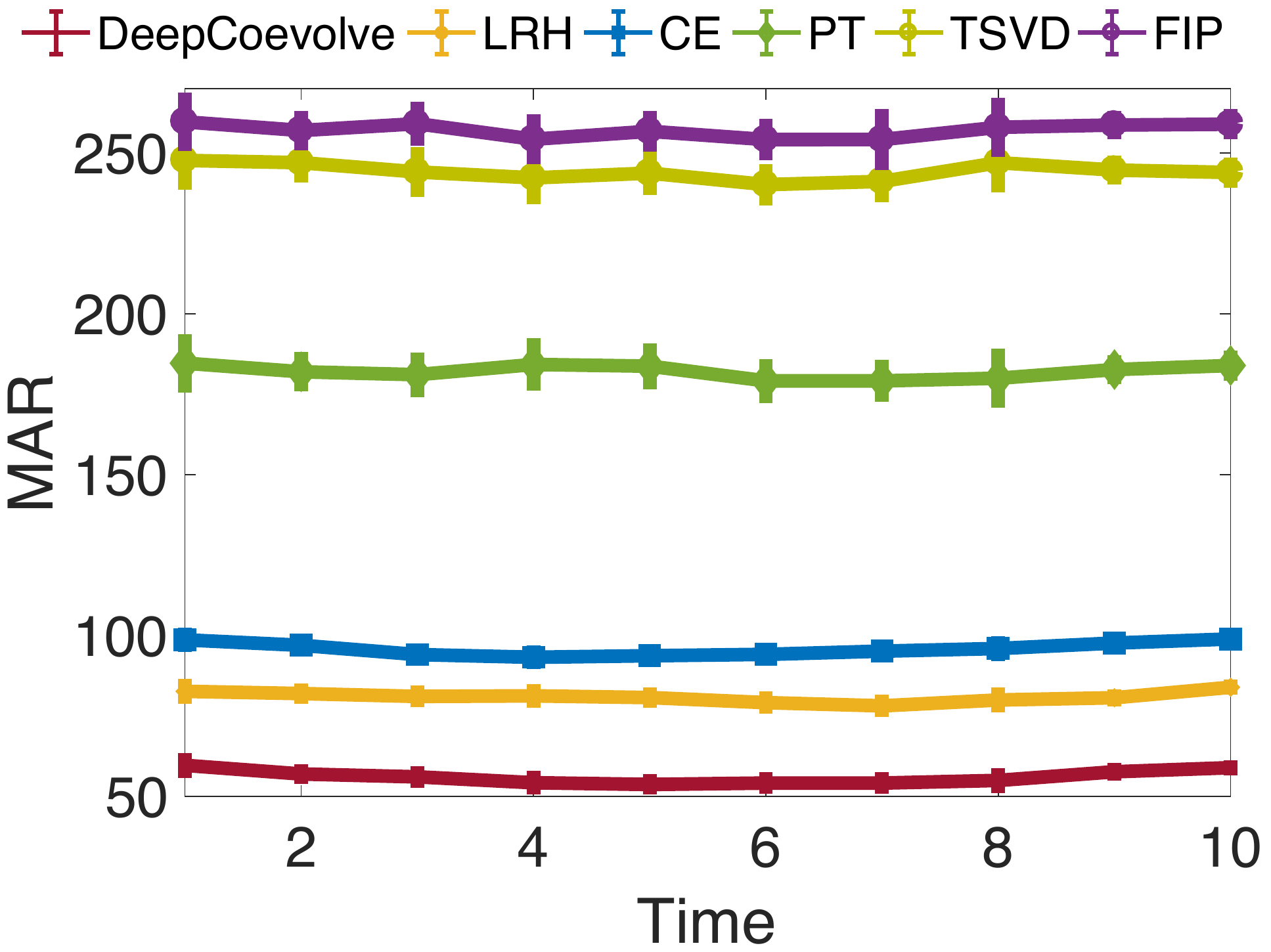}
& \includegraphics[width = 0.26\textwidth]{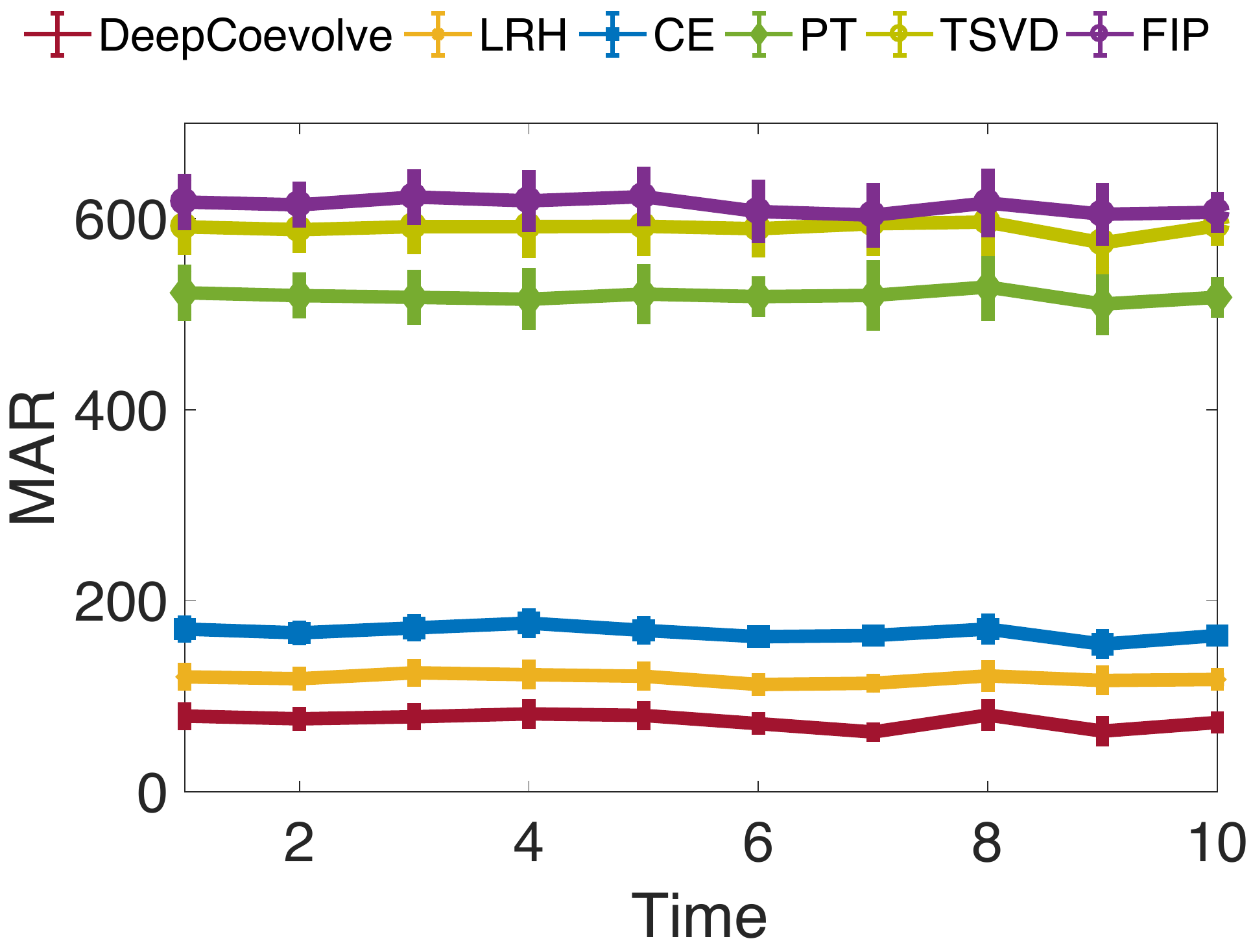}
& \includegraphics[width = 0.26\textwidth]{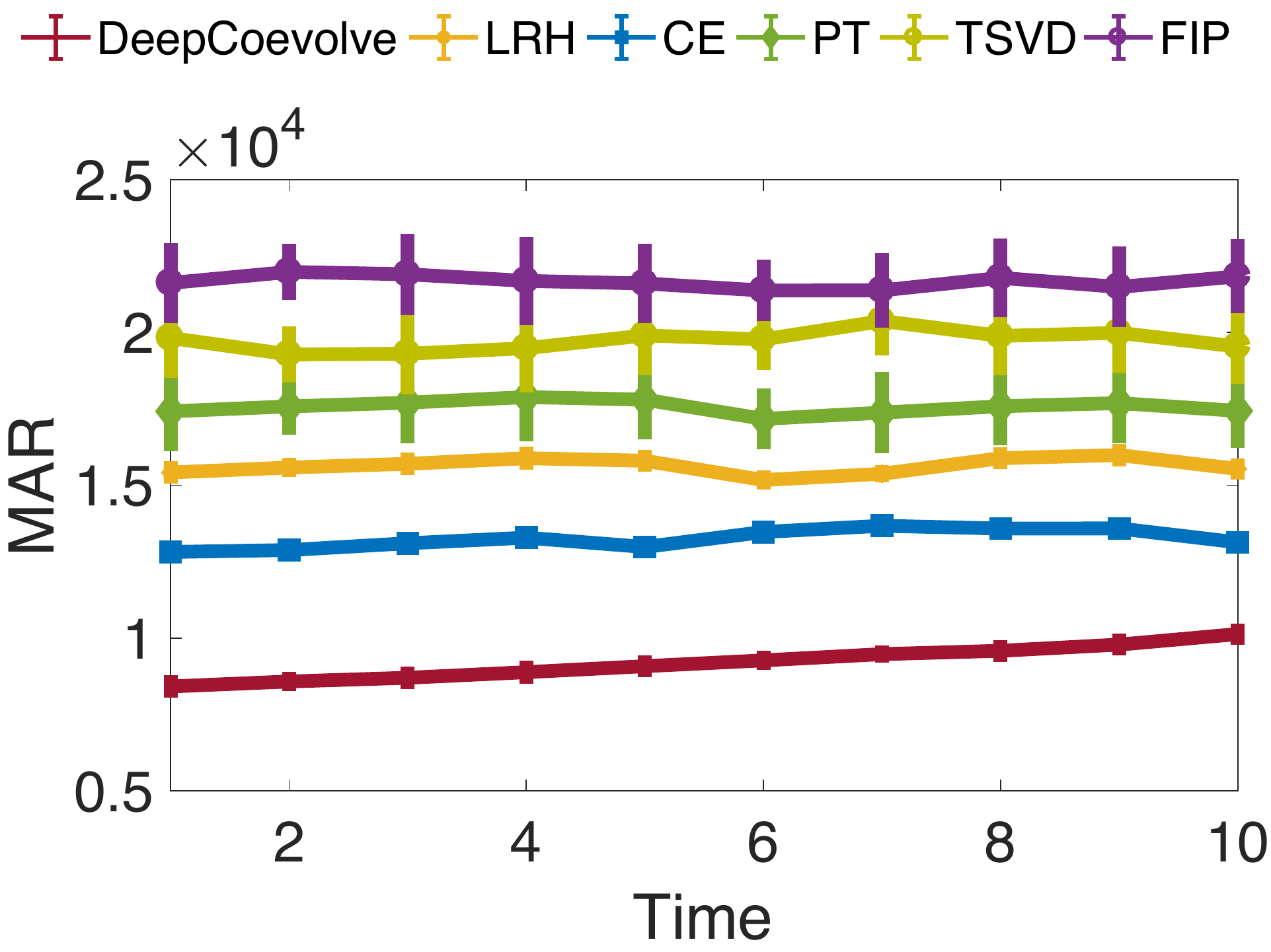}\\
& (a) IPTV	& (b) Reddit & (c) Yelp
\end{tabular}
\vspace{-4mm}
\caption{Item prediction over different time windows. We divide the entire test set into 10 bins, and report MAR in each bin.}
\label{fig:pred_windows}
\vspace{-3mm}
\end{figure*}

\begin{figure*}[t]
\small
\centering
\begin{tabular}{ccccc}
& \includegraphics[width = 0.25\textwidth]{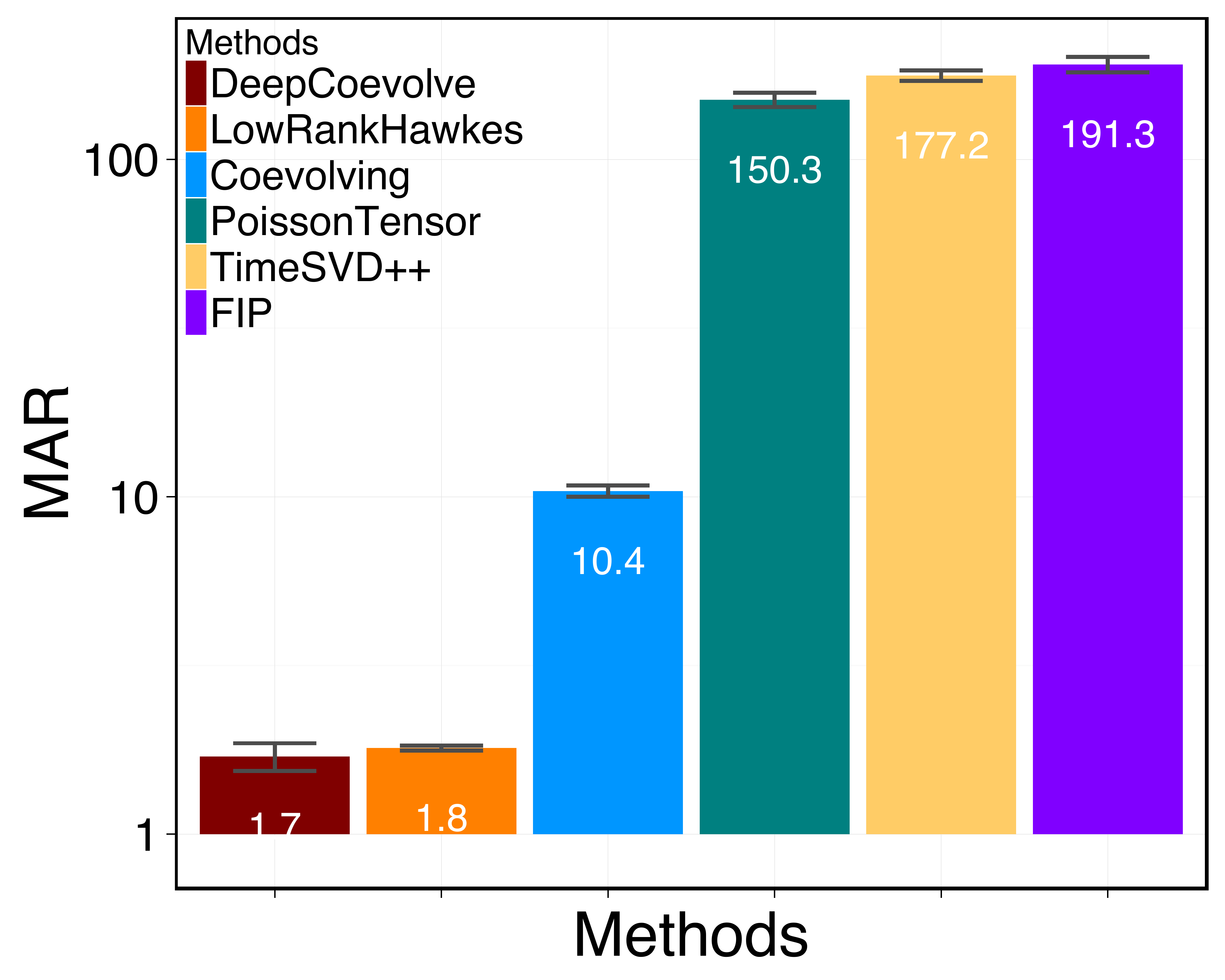}
& \includegraphics[width = 0.25\textwidth]{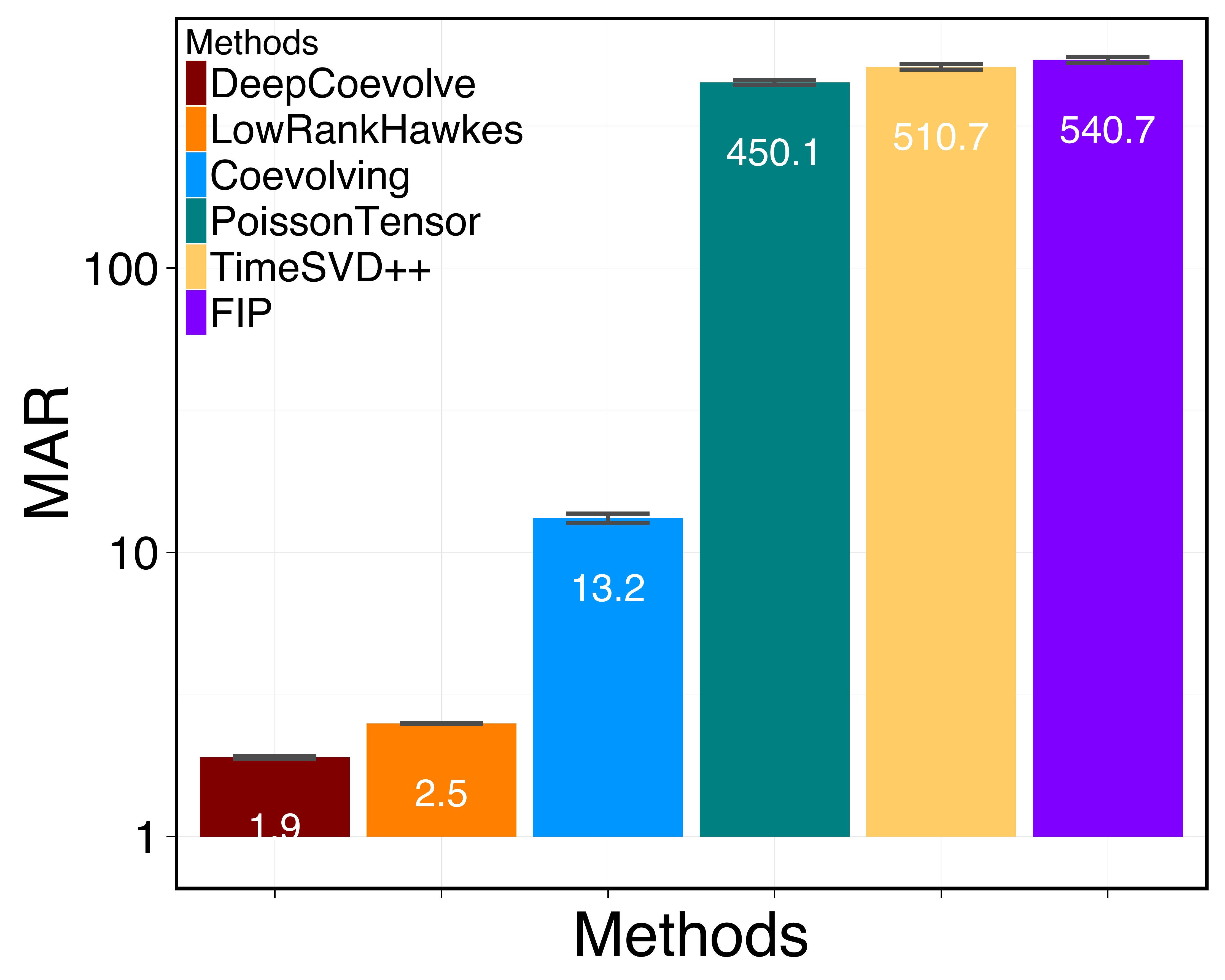}
& \includegraphics[width = 0.25\textwidth]{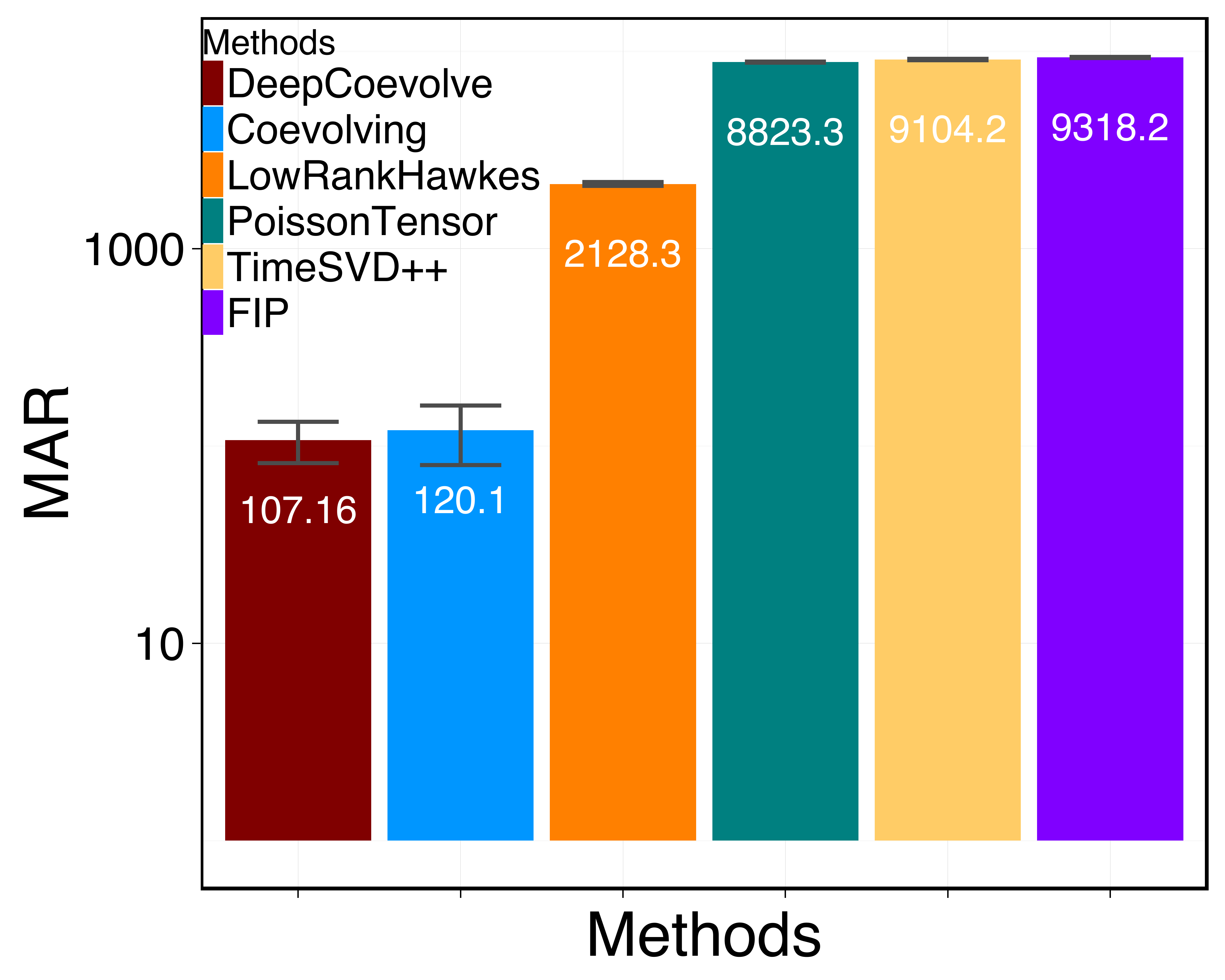}\\
& (a) IPTV	& (b) Reddit & (c) Yelp
\end{tabular}
\vspace{-3mm}
\caption{Item prediction performance on recurring events. }
\label{fig:pred_rec}
\vspace{-3mm}
\end{figure*}

\vspace{-3mm}
\subsection{Overall Performance Comparison}
\vspace{-0.5mm}

For each sequence of user activities, we use all the events up to time $T \cdot p$ as the training data, and the rest events as the testing data, where $T$ is the observation window.  We tune the latent rank of other baselines using 5-fold cross validation with grid search. We vary the proportion $p\in\cbr{0.7,0.72,0.74,0.76,0.78}$ and report the averaged results over five runs on two tasks (we will release code and data once published):

\begin{itemize}[leftmargin=*,nosep,nolistsep]
        \item \emph{Item prediction metric}. We report the Mean Average Rank (MAR) of each test item at the test time. Ideally, the item associated with the test time $t$ should rank one, hence smaller value indicates better predictive performance.
        \item \emph{Time prediction metric}. We report the Mean Absolute Error (MAE) between the predicted and true time. 
\end{itemize}

Figure~\ref{fig:exp} shows that \textsc{DeepCoevolve} significantly outperforms both epoch-based baselines and state-of-arts point process based methods.  \textsc{LowRankHawkes} has good performance on item prediction but not on time prediction, while \textsc{Coevolving} has good performance on time prediction but not on item prediction. We discuss the performance regarding these two metrics below.

\textbf{Item prediction.} 
Top row of Figure~\ref{fig:exp} shows the overall item recommendation performance across 5 different splits of datasets. Though the characteristics of datasets varies significantly, our proposed DeepCoevolve is doing consistently better than competitors. Note \textsc{LowRankHawkes} also achieves good item prediction performance, but not as good on the time prediction task. Since one only needs the rank of conditional density $p(\cdot)$ in \eq{eq:f} to conduct item prediction, \textsc{LowRankHawkes} may still be good at differentiating the conditional density function, but could not learn its actual value accurately, as shown in the time prediction task where the value of the conditional density function is needed for precise prediction. 

\textbf{Time prediction.} Figure~\ref{fig:exp} also shows that \textsc{DeepCoevolve} outperforms other methods. Compared with \textsc{LowRankHawkes},  our method has 6$\times$ improvement on Reddit, it has 10$\times$ improvement on Yelp, and 30$\times$ improvement on IPTV. The time unit is hour. Hence it has 2 weeks accuracy improvement on IPTV and 2 days on Reddit. This is important for online merchants to make time sensitive recommendations. An intuitive explanation is that our method accurately captures the \emph{nonlinear} pattern between user and item interactions. The competitor \textsc{LowRankHawkes} assumes specific parametric forms of the user-item interaction process, hence may not be accurate or expressive enough to capture real world temporal patterns. Furthermore, it models each user-item interaction dimension independently, which may lose the important affection from user's interaction with other items while predicting the current item's 
reoccurrence time. Our work also outperforms \textsc{Coevolving},~\eg, with around 3$\times$ MAE improve on IPTV. Moreover, the item prediction performance is also much better than \textsc{Coevolving}. It shows the importance of using RNN to capture the nonlinear embedding of user and item latent features, instead of the simple parametrized linear embedding in \textsc{Coevolving}.

\vspace{-1.5mm}
\subsection{Refined Performance Comparison}

{\bf Comparison in different time windows}. Figure~\ref{fig:pred_windows} further shows the item prediction performance on different time windows. Specifically, we divide the timeline of test set into 10 consecutive nonintersecting windows, and report MAR on each of them. Our DeepCoevolve is consistently better in different time periods. Moreover, all the point process based methods have stable performance with small variance. Since these models use new events to update the intensity function, the likelihood of unseen user-item interactions in the training set will have chance to get adjusted accordingly. 

\noindent
{\bf Performance on recurring events}. For the businesses like restaurants, it is also important to understand whether/when your customers will come back again. Here we compare the performance on those recurring events that appear both in training and testing. As expected in Figure~\ref{fig:pred_rec}, all the algorithms get better performance on this portion of events, compared to the overall results in Figure~\ref{fig:exp}. Moreover, the point process models benefit more from predicting recurring events. This justifies the fact that the more events we observe for a particular dimension (user-item pair), the better we can estimate its intensity and likelihood of future events. 


\vspace{-1.5mm}
\section{Related work} 

Recent work predominantly fix the latent features assigned to each user and item~\citep{SalMni08,ChePavCan09,AgaChe09,EksRieKon11,KorSil11,YanLonSmoEtal11,YiHonZhoLiu14,WanPal15}. In more sophisticated methods, the time is divided into epochs, and static latent feature models are applied to each epoch to capture some temporal aspects of the data~\citep{Koren09,KarAmaBalOli10,XioCheHuaTzu10,KarAmaBalOli10,XioCheHuaTzu10,ChiKol12,GulPai14,ChaRanMciBle15,BhaPhaZhoLee15,GopHofBle15,HidTik15,WanDonNeletal16}. For these methods, it is not clear how to choose the epoch length since different users may have very different timescale when they interact with items. Moreover, since the predictions are only in the resolution of the chosen epoch length, these methods typically are not good at time-sensitive question such as when a user will return to the item. A recent low-rank point process model~\citep{DuWanHeetal15} overcomes these limitations. However, it fails to capture the heterogeneous coevolutionary properties of 
user-item interactions. 

\begin{figure}[t]
\small
\centering
\includegraphics[scale=0.27]{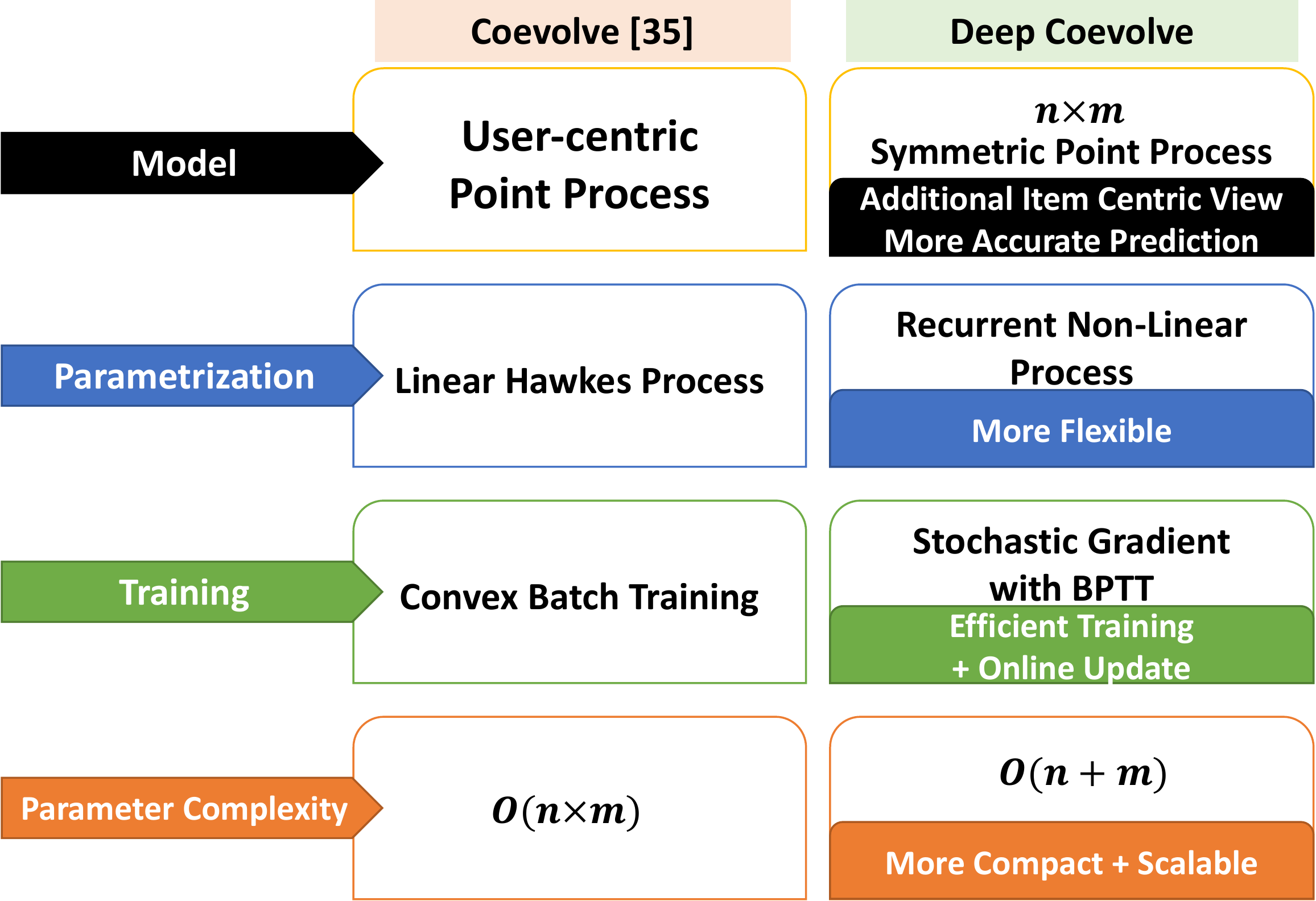}
\vspace{-2mm}
\caption{Comparison with the linear Hawkes model.}
\label{fig:cmp_ppt}
\vspace{-1mm}
\end{figure}

\citet{WanDuTriSon16} models the coevolutionary property, but uses a simple linear representation of the users' and items' latent features, which might not be expressive enough to capture the real world patterns. Our model is fundamentally different from~\cite{WanDuTriSon16}. We use RNN to model the complex dynamics of the feature embedding, which is more powerful due to the \emph{nonparametric} nature of our work and much improved prediction performance. More importantly, the model size is only \emph{linear} in the number of users and items, making our algorithm more scalable. Figure~\ref{fig:cmp_ppt} contains more details. 

As demonstrated in~\citet{DuDaiTriUpaGomSon16}, the nonlinear RNN unit is quite flexible to approximate many point process models. However, RMTPP is limited to learn only in one-dimensional point process setting. Our model is significantly different from RMTPP since we focus on the recommendation system setting with the idea of using multivariate point process and RNN to capture coevolutionary dynamics of latent features over a temporally evolving network. We further demonstrate that our model is very efficient
 even with the presence of RNN related parameters and can therefore be potentially applied to online setting.

In the deep learning community, very few work model the coevolution of users' and items' latent features and are still extensions of epoch based methods.~\citep{WanWanYue15} proposed a hierarchical Bayesian model that jointly performs learning for the content features and collaborative filtering for the ratings matrix.~\citep{HidKarBalTik16} applied RNN and adopt item-to-item recommendation approach with session based data.~\citep{TanXuLiu16} improved this model with techniques like data augmentation, temporal change adaptation. ~\citep{KoMayGro16} proposed collaborative RNN that extends collaborative filtering method to capture history of user behavior. Specifically, they used static global latent factors for items and assign separate latent factors for users that are dependent on their past history. ~\citep{SonElkHe16} extended the deep semantic structured model to capture multi-granularity temporal preference of users. They use separate RNN for each temporal granularity and combine them with feed forward 
network which models users' and items' long term static features. \cite{CheLiYanYu13} models the time change with piecewise constant function, but is not capable of predicting the future time point. Our work is unique in the sense that we explicitly treat time stamps as random variables and model the coevolution of users' and items' latent features using temporal point processes and deep learning model over evolving graph.

\section{Conclusions}
We proposed an expressive and efficient framework to model the \emph{nonlinear} coevolution nature of users' and items' embeddings. Moreover, the user and item's evolving and coevolving processes are captured by the RNN.  We further developed an efficient stochastic training algorithm for the coevolving user-item netowrks. We demonstrate the superior performance of our method on both the time and item prediction task, which is not possible by most prior work. Future work includes extending to other social applications, such as group dynamics in message services. 



\end{document}